\documentclass[runningheads]{llncs}
\usepackage[T1]{fontenc}

\usepackage{graphicx}

\usepackage{hyperref}       
\usepackage{color}

\usepackage{url}            
\usepackage{booktabs}       
\usepackage{amsfonts}       
\usepackage{nicefrac}       
\usepackage{xfrac}

\usepackage{tikz}

\usepackage{microtype}      
\usepackage{bigstrut}
\usepackage{xcolor}
\usepackage{colortbl}
\usepackage{hhline}
\usepackage{diagbox}

\usepackage[shortlabels]{enumitem}
\usepackage{breakcites}  

\usepackage{amssymb}
\usepackage{amsmath}        
\usepackage{siunitx}
\usepackage{array,multirow}

\usepackage{colortbl} 
\usepackage{multirow}
\usepackage{bigstrut}
\usepackage{diagbox}

\usepackage{xargs} 

\newtheorem{prop}[theorem]{Proposition}
\newtheorem{rem}[theorem]{Remark}

\newcommand{\R}{\mathbb{R}}
\newcommand{\N}{\mathbb{N}}
\newcommand{\Z}{\mathbb{Z}}
\newcommand{\calO}{\mathcal{O}}

\newcommand{\SO}{\mathrm{SO}}
\newcommand{\Hm}{\mathrm{RH}}

\newcommand{\PY}{\mathrm{PY}}

\newcommand{\abs}[1]{\left\vert #1 \right\vert}

\newcommand{\sse}{\subseteq}
\newcommand{\sph}{\mathbb{S}}
\newcommand{\id}{\mathrm{id}}
\newcommand{\dd}{\mathrm{d}}

\newcommand{\h}{\mathrm{h}}

\newcommand{\ED}{EfficientDet}
\newcommand{\EP}{EfficientPose}
\newcommand{\LM}{LINEMOD}
\newcommand{\LMO}{Occlusion LINEMOD}

\newcommand{\add}{\textsc{add(-s)-$0.1$d}}

\newcommand{\result}[1]{$#1\%$}
\newcommand{\resultbf}[1]{$\mathbf{#1\%}$}

\newcommand{\newcorrectcell}{}

\usetikzlibrary{calc, positioning, shapes, arrows}
\tikzstyle{3x3} = [rectangle, on grid, draw=green, text=black, ultra thick, rotate=90, minimum width=1.9cm]
\tikzstyle{3x3s2} = [rectangle, dotted, on grid, draw=green, text=black, ultra thick, rotate=90, minimum width=1.9cm] 
\tikzstyle{1x1} = [rectangle, on grid, draw=blue, text=black, ultra thick, rotate=90, minimum width=1.9cm]

\newcommand{\ugreendotted}[1]{%
    \protect\tikz[baseline=(todotted.base)]{
        \protect\node[inner sep=1pt,outer sep=0pt] (todotted) {#1};
        \protect\draw[dotted, draw=green, ultra thick] (todotted.south west) -- (todotted.south east);
    }%
}%

\begin{document}

\title{Rigidity Preserving Image Transformations and Equivariance in Perspective \thanks{Brynte and Bökman contributed equally to this work.}}

\author{
Lucas Brynte\inst{1} \and Georg Bökman\inst{1} \and Axel Flinth\inst{2} \and Fredrik Kahl\inst{1}
}

\institute{Department of Electrical Engineering, Chalmers University of Technology
\and
Department of Mathematics and Mathematical Statistics, Umeå University}

\maketitle

\begin{abstract}
  We characterize the class of image plane transformations which realize rigid camera motions and call these transformations `rigidity preserving'. It turns out that the only rigidity preserving image transformations are homographies corresponding to rotating the camera. In particular, 2D translations of pinhole images are not rigidity preserving. Hence, when using CNNs for 3D inference tasks, it can be beneficial to modify the inductive bias from equivariance 
  w.r.t.
  translations to equivariance 
  w.r.t. 
  rotational homographies.
  We investigate how equivariance with respect to 
  rotational homographies
  can be approximated in CNNs, and test our ideas on 
  6D object pose estimation. Experimentally, we improve on a competitive baseline.
\end{abstract}

\begin{figure}[ht]
    \centering
    \includegraphics[width=.95\columnwidth,trim=.3cm .3cm .3cm .3cm,clip=true]{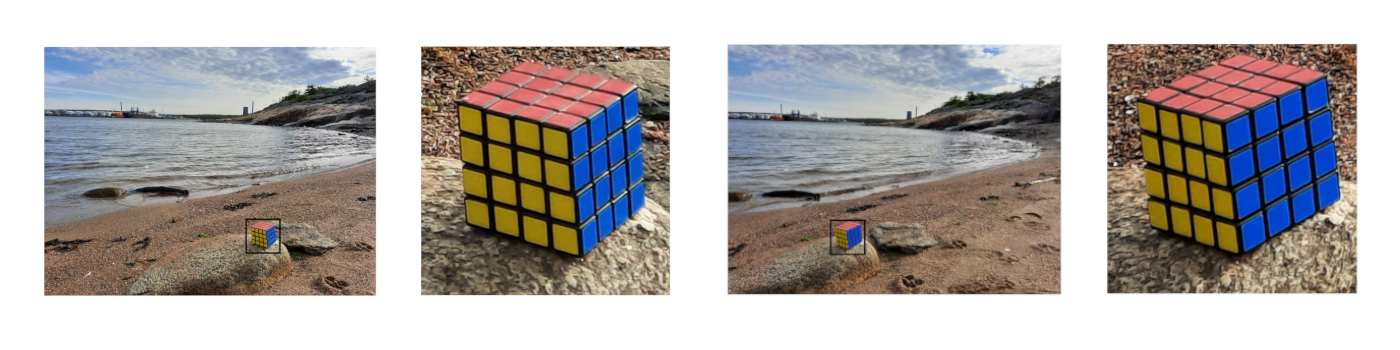}
    \caption{Two pictures taken by the same camera, but with camera positions translated relative to each other. Notably, the  depictions of an object in the two images are \emph{not} translations of each other. 
    In this paper, we show that given a single image and a rigid camera motion that is not a pure rotation, there is no way to unambiguously determine the second image -- they are not related by a \emph{rigidity preserving image transformation}.
    \label{fig:transldemo}}
\end{figure}


\section{Introduction}
An important property of convolution operators (e.g., linear convolutional layers in neural networks) is their \emph{translation equivariance}, i.e.\ that translation and application of the network commute. Put in slightly formal terms, we say that an operator $\Psi$ is translation equivariant if for each input $f$ and each translation $t$, we have
\begin{equation} \nonumber
    \Psi ( \Lambda_t f) = \Lambda_t (\Psi f),
\end{equation}
where $\Lambda_tf(x) = f(x-t)$. 
Through a global average pooling, this equivariance can easily be turned in to an \emph{invariance} as well, i.e.\ $\widehat{\Psi}(f)=\widehat{\Psi}(\Lambda_t f)$. It is clear that such properties might be beneficial when designing neural networks. To be concrete, consider a network designed for detecting the presence of a cat in an image. That network should react similarly to a cat in the lower right corner as to one the upper right corner -- it should be invariant to translations. If the network instead should detect the position of the cat, the prediction should translate with the cat -- this is equivariance.

However, as we show in this paper, a translation of the camera does not correspond to translation of the image for any non-planar surface patch in the scene. See Figure~\ref{fig:transldemo} for an example.
One
can even prove that translations of the image do not correspond to images taken following any rigid camera movements.
{
While such rigid motion can under certain circumstances and to some extent be approximated by image translation,
we argue for considering
a different set of image transformations,
that actually correspond to rigid camera motion. 
}

We refer to such image transformations as
\emph{rigidity preserving}. 
In Section~\ref{sec:natural_image_transformations}, we provide a formal definition, and show that these transformations exactly consist of \emph{rotational homographies}.

\subsection{Related work}
Convolutional layers have been a workhorse of neural networks for several decades \cite{fukushima-1980,krizhevsky-nips-2012,lecun-1989}.
Even though convolution operators yield translation equivariance, the whole network is typically not equivariant due to  subsampling  and padding operations \cite{Kayhan_2020_CVPR,Islam*2020How}. Attempts to redesign modern CNNs to achieve translation equivariance have resulted in improved robustness, but degraded performance \cite{azulay-weiss-jmlr-2019,zhang2019shiftinvar}. In this paper, further evidence that translation equivariance may not be the ideal objective for perspective images is provided.

Our work is also related to spherical image representations, which are frequently used for wide-angle cameras, e.g., fish-eye cameras \cite{matsuki2018omnidirectional}. Recently, several frameworks for obtaining $\SO(3)$-equivariance have been developed \cite{cohen2019gauge,cohen-etal-iclr-2018,defferrard2019deepsphere,esteves-nips-2020,esteves2018learning,jiang2019spherical-short}.
However, ordinary perspective images, which can be mapped to the sphere, typically have a well-defined orientation, so $\SO(3)$-equivariance is not necessarily desired. 
Mapping a perspective image to the sphere also yields a signal
only
on a small part of the sphere, which may severely limit performance for methods designed for
signals
on the entire sphere.
A further complicating factor is that many of these approaches work in the spherical Fourier domain. 
We will work in the spatial domain and use some of the invariant properties of the sphere representation. From this viewpoint, our approach resembles methods
that adapt
standard CNNs to be applicable to spherical images \cite{coors2018spherenet,eder2020tangent,su2019kernel},
though this is not our aim.

Our pitch-yaw-group equivariant network construction is related to so called group convolutional neural networks  \cite{cohen2016group,lang2020wigner,yarotsky2021universal}.
In our setting, the approach boils down to applying a convolution kernel in a coordinate system defined by the azimuthal equidistant map projection. Another work drawing inspiration from map projections to define sampling schemes on the sphere is \cite{eder2019convolutions}. More information in general on map projections can be found in \cite{snyder-1987}. 

Data augmentation is a standard technique for increasing the amount of image data with slightly modified images. Various geometric transformations are used in the literature. For instance, in \cite{simonyan-arxiv-2014} cropping provides randomly translated image patches, in \cite{bukschat2020efficientpose} scalings and rotations in the image plane are applied, and in \cite{li2020hierarchical} 2D affine transformations.
More elaborate transformations, such as Möbius transformations \cite{zhou2020data} have also been investigated with improved performance and generalization as a result. More specifically, 2D homography augmentations are used in \cite{wang_perspective_2020} for object detection.
An augmented image is obtained by transforming the corners of the original image to four random points.
In contrast to our approach, this procedure
does not necessarily result
in an image possibly taken by the original camera.
Systematic approaches for evaluating various data augmentation techniques can be found in, e.g., \cite{cubuk2019autoaugment,cubuk-nips-2020,zhang-icml-2020}.

For neural networks that estimate 3D properties from images, e.g., 6D object pose estimation \cite{bukschat2020efficientpose,Hu_2020_CVPR,liu-ral-2021} and visual localization \cite{brachmann-pami-2021,li2020hierarchical,sarlin2021feature}, data augmentation is a delicate matter since
while the input image is transformed, the 3D target also has to undergo a corresponding
transformation. As an example, in 6D object pose estimation, an image rotation around the principal point corresponds to a 3D rotation around the optical axis of the object pose, cf.\ \cite{bukschat2020efficientpose}.

In \cite{mahendran_3d_2017}, the
pose
of objects in the images are rotated, and the resulting transformation of the image is approximated with a homography. This approach does not necessarily result in an image that could have been taken by the original camera. The type of augmentations we are considering in this article appears to only have been used in~\cite{Cheng_2021_IEEETransInstrumMeas}, which is currently the SOTA method for the \LM{} dataset. In this article, we consolidate this approach by our theoretical considerations.

In \cite{wang-mscthesis-2021}, different geometric augmentation strategies are experimentally evaluated for pose estimation and it is found that some transformations (scaling and 2D translation) do not correspond to any rigid camera motion and may degrade
performance, while 2D rotation around the principal point is geometrically sound, and enhances performance in their experiments.
This is consistent with our theoretical findings - the group of rigidity preserving image transformations is a subset of the homographies, and includes 2D rotation as a special case.

\subsection{Contributions}

CNNs are very good at learning from image data, but their inductive bias -- translation equivariance -- is not perfectly suited to 3D tasks due to perspective effects.
This fact is implicitly understood by the community, but we believe that it has not been laid out in a precise manner in earlier work.

We make a precise statement in Theorem~\ref{th:characterization2}. It shows
that, under the assumption of unknown depth, no rigid camera motions except pure rotations yield corresponding 
transformations of the image plane -- neither a homography, nor an image translation, nor any, potentially more obscure, transformation. 
Of course, the pure camera rotations do yield such transformations, namely rotational homographies.

It follows from Theorem~\ref{th:characterization2} that image translations (the equivariance group of CNNs) 
do not correspond to any camera motions.
This motivates
a shifted focus to
equivariance w.r.t.
rotational homographies, as they \emph{do} correspond to camera motions.

We propose two simple methods
for approximating such equivariance in CNNs:
In Section~\ref{sec:rotequiv}, the $\mathrm{PY}$-transformation, 
which is a novel approach to learning on images and in Section~\ref{sec:aug}, data augmentation with rotational homographies, which has only been used scarcely in the prior literature.
Finally, in Section~\ref{sec:exp} we experimentally show that these two methods improve the performance of EfficientPose \cite{bukschat2020efficientpose} on 6D object pose estimation.

\paragraph{Limitations. }
Although the aim of our work is to boost the performance of computer vision systems, the expected performance boost is dependent on whether a camera has a wide or narrow field of view: Near the principal point, the translation equivariance of a regular CNN model
already approximates
rigid camera motion quite well.

In the other extreme, while the pitch-yaw group (defined in Section \ref{ssec:PY}) can be applied to spherical images, it should not be seen as a new way of designing equivariant networks on the
full sphere.
Very far away from the north pole it too creates large distortions, despite not being as severe as a regular perspective projection in general.

Experiments on visual localization, which we present in Section~\ref{sec:appendix_localization} of the supplementary material, indicate that the gain of improving the rotation equivariance of a CNN is not always as clear cut as in our main object pose experiments.

\section{Rigidity preserving image transformations}\label{sec:natural_image_transformations}
We assume images acquired with a pinhole camera with \emph{known intrinsic calibration}. Without loss of generality, we
choose an orthonormal coordinate system $(e_0,e_1,e_2)$ in $\R^3$
with the camera center at the origin, the optical axis along the $x_2$-axis, and the image plane at $x_2=1$,
so that a non-occluded 3D point at position $x=\begin{bmatrix}x_0,x_1,x_2 \end{bmatrix}^T\in \R^3$ with $x_2 >0$
projects
to an image point at position
$$\pi(x) = \begin{bmatrix}x_0/x_2, x_1/x_2\end{bmatrix}^T \in \R^2 .
$$
To simplify notation, we may think of this as a point
in the \emph{projective plane} $\mathbb{P}^2$, so that
$\pi(x) = \begin{bmatrix}x_0,x_1,x_2\end{bmatrix}^T$
in homogeneous coordinates. We model images as functions on $\mathbb{P}^2$.

{
Let us consider an image acquired by taking a picture of a scene. Mathematically, this corresponds to projecting the three-dimensional scene to some set $B\sse \mathbb{P}^2$. Importantly, we assume that the 3D shape and depth of the scene is unknown to us.

If we move the scene (or by duality, the camera) using a rigid transformation, the pixels in the image representing the
scene
will also be transformed via some map.
}
 Let us name such transformations.

\begin{definition}
    Let $B \sse \mathbb{P}^2$ be a set containing at least two points.
    A map $\eta$ is a
    \emph{rigidity preserving image transformation on $B$}
    if there exists a rigid transformation $\rho$ of $\R^3$ so that
    \begin{align} \label{eq:eqmappings}
        \eta
        \circ \pi(x) = \pi \circ \rho(x),  \quad \mbox{for all }
        x\in \pi^{-1}(B).
        \end{align}
\end{definition}
\begin{rem}
The unknown depth assumption is manifested in requiring \eqref{eq:eqmappings} to hold for all $x \in \pi^{-1}(B)$, and not only for a specific scene $U$ with $\pi(U)=B$. Note that if depth information is available, one can always construct $\eta$ by rendering the scene at two different camera locations (provided no occlusions occur).
\end{rem}

A mapping $\mathbb{P}^2 \to \mathbb{P}^2$ is called a {\bf homography} if it can be expressed as $x \mapsto y=Hx$ where $H$ is a non-singular $3 \times 3$ matrix and $x,y$ are $3$-vectors of homogeneous coordinates representing the points in $\mathbb{P}^2$. It turns out that the rigidity preserving image transformations are a subset of the homographies, namely the \emph{rotational homographies}.

\begin{definition}
    A homography $H$ is a \emph{rotational homography} if the matrix representing it is an element of $\SO(3)$. We denote the set of rotational homographies by $\Hm$.
\end{definition}

This sets us up for our main theorem.
\begin{theorem} \label{th:characterization2} Rigidity preserving image transformations correspond exactly to the rotational homographies $\Hm$. More precisely:
    \begin{enumerate}[(i)]
    \item
    If $\eta$ is a rigidity preserving image transformation on $B$, it is a rotational homography on $B$.
    \item Every $R \in \SO(3)$ gives rise to a rigidity preserving image transformation $\eta$, where the rigid transformation $\rho$ in \eqref{eq:eqmappings} is given by $\rho(x) = Rx$.
    \end{enumerate}
\end{theorem}
\begin{proof}
What needs to be shown is that the rigid transformation $\rho$ in \eqref{eq:eqmappings} must be a pure rotation, that is, $\rho(x) = Rx$ for some $R \in \SO(3)$. Towards a contradiction, assume that  $\rho(x) = Rx + t$, with $R\in\SO(3)$ and $t \not= 0 \in\mathbb R^3$. Since $B$ by assumption contains at least two points,
there exists a point 
$u\in \pi^{-1}(B)$
such that $Ru$ and $t$ are linearly independent. Now, let  $\lambda\neq 1$. We  have $\pi(u)=\pi(\lambda u)$ and hence $\eta(\pi(u))=\eta(\pi(\lambda u))$ for our rigidity preserving image transformation $\eta$. This, together with \eqref{eq:eqmappings}, implies that $\pi(\rho(u)) = \pi(\rho(\lambda u))$, i.e. that $\rho(u)$ and $\rho(\lambda u)$ are collinear. However, their cross-product is non-zero:
    \[
       \rho(u)\times\rho(\lambda u) = (Ru+t)\times(\lambda Ru+t) = (1-\lambda) (Ru)\times t,
    \]
    which leads to a contradiction, and hence, we can conclude that $t=0$.
    
(ii) Define $\eta$ by $\eta(y) = Ry$, where $y$ is a 3-vector of homogeneous coordinates. $\eta$ is well defined since for every $\lambda\neq0$ we have $\eta(\lambda y) = \lambda \eta(y)$ so that every homogeneous representation of $y$ is mapped to a homogeneous representation of 
$\eta(y)$. It is clear that $\eta$ satisfies \eqref{eq:eqmappings} with $\rho(x)=Rx$.

\end{proof}

\begin{rem} \label{rem:translhom}
\phantom{Hej} 
\begin{itemize} 
\item A careful inspection of the proof shows that the theorem holds already when \eqref{eq:eqmappings} holds on some open subset of $\pi^{-1}(B)$. This relaxed assumption can be interpreted as not having \emph{exact} depth information.
\item A translation in the image plane by a vector $\tau=\begin{bmatrix} \tau_0 & \tau_1\end{bmatrix}^T$
can be written as a homography $H$ with 
$$ H =\begin{bmatrix} I_{2 \times 2} & \tau  \\  0 & 1
\end{bmatrix}.$$
Note, however, that this never corresponds to a rigidity preserving image transformation as $H$ is not an orthogonal matrix. \label{rem:homography}
\end{itemize}
\end{rem}

The above proof relies heavily on the assumption of unknown depth. How much more can be done when the depth is known? Are the translations rigidity preserving then? The following proposition shows that unless we are
taking an image of a 3D plane
parallel to the image plane, a rigid camera motion can only be equivalent to an image  translation on singular sets.

\begin{prop}\label{prop:singularSets}
Let $\tau$ be a translation in the image plane not equal to the identity. Suppose that for some rigid motion $\rho$ and some set $U \sse \R^3 \backslash\{0\}$, $\tau \circ \pi =\pi \circ \rho$ on $U$. Then $U$ is a union of a continuous curve $\gamma : \R\backslash M\to \R^3$ (where $M$ is a set of at most three points), and a union of either
\begin{itemize}
    \item a plane parallel to $\mathrm{span}(e_0,e_1)$, and possibly one more line.
    \item between one and three lines.
\end{itemize} 
\end{prop}

The proof can be found in Section~\ref{sec:proof_singularSets} of the supplementary material. Its main idea is to reinterpret the equation  $\tau(\pi(x))=\pi(\rho(x))$ as a generalized eigenvalue problem, and then analyze the solutions to that problem.

Theorem~\ref{th:characterization2} and Proposition~\ref{prop:singularSets} should convince the reader that translations are not the most natural transformations of
perspective
images, and that the rotational homographies $\Hm$ are more suited. We will explore two
approaches for handling equivariance
w.r.t.
rigidity preserving image transformations in the next two sections. The first one is to develop neural networks that are equivariant by design. The second approach is to learn equivariance by data augmentation.

\section{Rotational homography equivariance}\label{sec:rotequiv}
Theorem \ref{th:characterization2} suggests that the natural equivariance for neural networks acting on images is \emph{not} the group of translations, but rather the rotational homography group.
Accordingly we should try to design networks that are equivariant with respect to this action. The following proposition shows that a straightforward relation between $\Hm$ and $\SO(3)$ gives us a simple way to do this.

\begin{prop} \label{prop:HvsSO}
(i) The rotational homographies $\Hm$ form a subgroup of the group of diffeomorphisms on $\mathbb{P}^2$. The subgroup is isomorphic to $\SO(3)$ equipped with matrix multiplication.

(ii) Let $\psi$ be an $\SO(3)$-equivariant network. Then
    \begin{align*}
        \phi: f \mapsto (\psi (f \circ \pi))\circ \pi^{-1}
    \end{align*}
    is an $\Hm$-equivariant network. Here, $\pi^{-1}$ maps into the upper half sphere.
\end{prop}

The proofs can be found in
Section~\ref{prop:proof_HvsSO} of the supplementary material.
The result gives a direct way to design an $\Hm$-equivariant network: 
we transform the input image $f$ to the spherical image $f \circ \pi $. Applying the $\SO(3)$-equivariant network  $\psi$ to $f\circ \pi$ is then exactly equivalent to applying the $\Hm$-equivariant network $\phi$ to $f$!

\subsection{The pitch-yaw group}\label{ssec:PY}
There is a caveat to using a network with invariance/equivariance with respect to the entirety of $\Hm$. The group $\Hm$ namely includes rotations \emph{in the image plane} -- these correspond exactly to the $h \in \Hm$ associated to $R$ around the $x_2$-axis. This behavior can sometimes be problematic. A somewhat contrived, but still relevant example, is given by a self-driving car. Full $\Hm$-invariance would mean that it cannot distinguish between a speed limit of $90$ km/h and one of $06$ km/h!

Hence, it can be beneficial to design a network which is only equivariant with respect to rotations around axes in the plane $\mathrm{span}(e_0,e_1)$.
However, we have the following fact.

\begin{prop} \label{prop:pitchyawnegative}
    No proper subgroup of $\SO(3)$ contains both the rotations around the $x_0$-axis (pitches) and the rotations around the $x_1$-axis (yaws).
\end{prop}

\begin{proof}
   If $G$ is a subgroup containing pitches and yaws, it must contain all matrices of the form
    \begin{align*}
        R= R_{x_0}(\alpha) R_{x_1}(\beta) R_{x_0}(\gamma),
    \end{align*}
    where $R_{x_0}(\alpha)$ and $R_{x_0}(\gamma)$ are rotations about the ${x_0}$-axis (i.e. pitches) and $R_{x_1}(\beta)$ a rotation about the ${x_1}$-axis (i.e. a yaw).  A classical result of differential geometry (the existence of \emph{Euler angles}) however states that every rotation $R$ can be written in that form. Hence, such a subgroup cannot be proper.  
\end{proof}
Proposition~\ref{prop:pitchyawnegative} prohibits us from using any form of restriction of the action of $\SO(3)$ on the sphere to design a network only equivariant to pitches and yaws. To get an idea of what to do instead, let us notice that any rotation around an axis in the $\mathrm{span}(e_0,e_1)$-plane can be written as $\exp(\alpha_0 C_0 + \alpha_1 C_1)\sse \SO(3)$, where $\exp$ is the matrix exponential, and $C_0$ and $C_1$ are the skew-symmetric matrices corresponding to the linear maps $v \mapsto e_0 \times v$ and $v \mapsto e_1 \times v$, respectively. Such rotations will in the following be referred to as pitch-yaws. If we look at the ``logarithms'' $\alpha_0 C_0 +\alpha_1 C_1$ directly, we can define a group structure by addition on this set. We identify $\alpha_0 C_0 +\alpha_1 C_1$ with the vector $\alpha \in \R^2$.

\begin{definition}
    We define $\PY$ as $\mathrm{span}(C_0,C_1)$, and equip it with the group structure induced by vector addition.
\end{definition}

Given two functions $F$ and $G$ on $\PY$, we may now define the group-convolution between them via
\begin{align} \label{eq:PYconvolution}
    F*G(\alpha) = \int_{\PY} F(\alpha-\beta) G(\beta) \dd \beta, \quad \alpha \in \PY.
\end{align}
Using such filter convolutions as linear layers in a network induces $\PY$-equivariance.

Our spherical images 
$\hat{f}$
are functions on the sphere, and not on $\PY$. 
We may however convert them into functions $F$ on $\PY$ via setting $F(\alpha) = \hat{f}(\exp(\alpha)e_2)$ for $\abs{\alpha}<\pi$, and $F(\alpha)=0$ else.
In this way, we may convolve spherical images with filters on $\PY$. Practically, this means that by sampling the image in points $\{\exp(\alpha)e_2 \, \vert \, \alpha \in G, \ \abs{\alpha}<\pi \}$,
 where $G \sse \PY$ is a regular grid, and applying a standard CNN to the resampled image, we obtain a network which is $\PY$-equivariant.

\begin{figure*}[t]
    \centering
    \includegraphics[width=.88\textwidth]{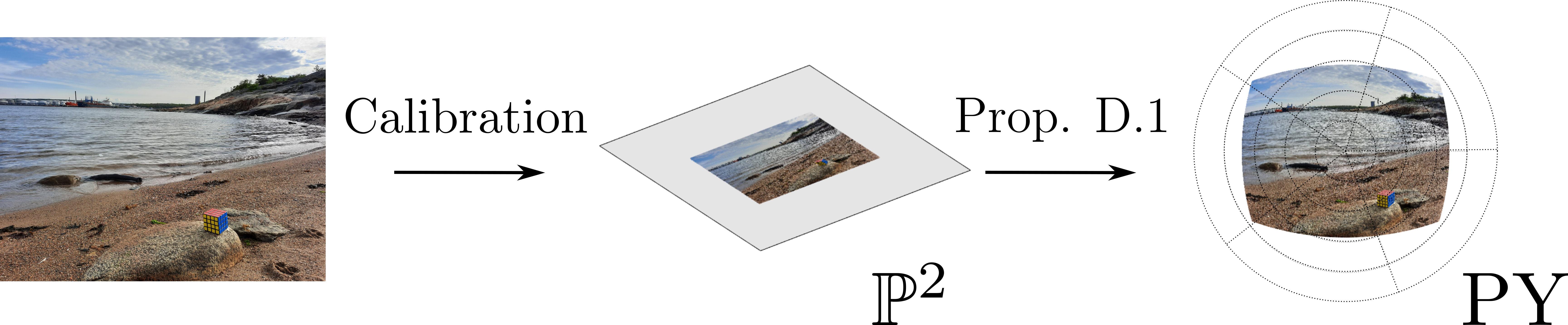}
    \caption{Transforming (warping) a $\mathbb{P}^2$-image represented in the pixel domain to a $\PY$-image in the pixel domain. Prop.~\ref{prop:warp}, which is contained in the supplementary material, gives an explicit formula for the map from $\mathbb P^2$ to $\mathrm{PY}$. }
    \label{fig:warp}
\end{figure*}
This resampling is,  up to a shift of the polar angle, to the \emph{azimuthal equidistant map projection} \cite{snyder-1987}. This projection maps a point with spherical coordinates $(\vartheta,\varphi)$ on the sphere to the point with radius $\vartheta$ and polar angle $\varphi$ in the plane.  Similarly, we may characterize the mapping of an image on $\mathbb{P}^2$ to an image on $\PY$ as a simple radial transformation.
{An example of a warped image is provided in Figure~\ref{fig:warp}.}
More information 
is provided
 in Section~\ref{sec:proofs_warp} in the supplementary material.

The azimuthal equidistant map projection is just one of many that can be used for mapping the sphere to a plane. Hence, there are many more strategies of resampling a spherical image and convolution (see for instance \cite{boomsma2017spherical,eder2019convolutions,coors2018spherenet}).
We leave it for future work to analyze other resampling strategies.
\begin{rem} 
In Section~\ref{sec:groupaction} of the supplementary material, we discuss how this construction can be interpreted as defining an \emph{action} $\chi_\alpha$ of $\PY$ on a countable union of distinct spheres $\N \times \mathbb{S}^2$ (up to a set of measure zero). By viewing the images as functions on one of these spheres, we may convolve them with filters on $\PY$ directly.

\end{rem}

\begin{figure*}[t]
    \centering
    \includegraphics[width=.88\textwidth]{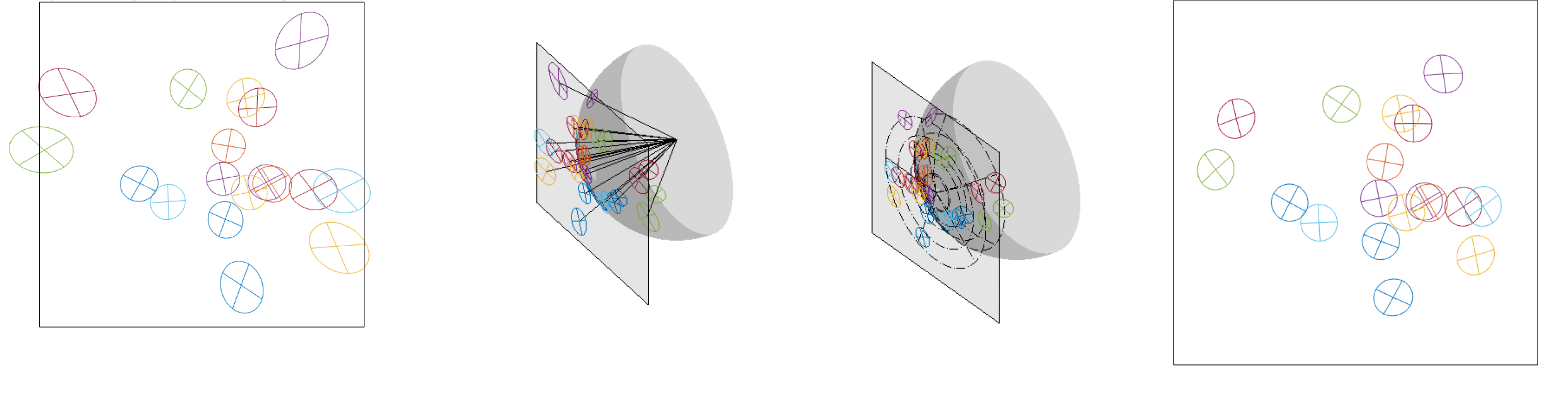}  
    \caption{The two image representations. The pinhole camera model and the conventional image representation on $\mathbb{P}^2$ to the left, and our proposed pitch-yaw model and image on $\PY$ to the right, corresponding to the azimuthal equidistant map projection.
    Shown are the projections of identical shapes placed at equal distance from the camera but at different angles from the principal axis.}
    \label{fig:sphere_to_circle}
\end{figure*}
\subsection{The approximating property of \texorpdfstring{$\PY$}{PY} } \label{sec:Appr}
In Figure \ref{fig:sphere_to_circle}, we show an example of a spherical image being represented on $\mathbb{P}^2$ and on $\PY$. The identical small spherical caps are clearly distorted by the projections onto $\mathbb{P}^2$ and  $\PY$, respectively. These distortions are however apparently less severe for the $\PY$-representation. 

We can analyze this via comparing the actions of $\PY$ and the corresponding elements in $\exp(\PY)$, i.e. comparing $\PY$-translating $\theta$ with $\alpha$ to rotating $\theta$ by the element $\exp(\alpha) \in \SO(3)$. 
If the magnitude of $\alpha$ is small, and $\theta$ is close to the north pole $e_2$, the two actions are very similar. Note that while this is not enough when full spherical images are concerned, it is acceptable in our setting:
Unless using very wide lenses,
images will be supported close enough to the principal point.

\begin{figure*}[t]
    \centering
    \includegraphics[width=.89\textwidth]{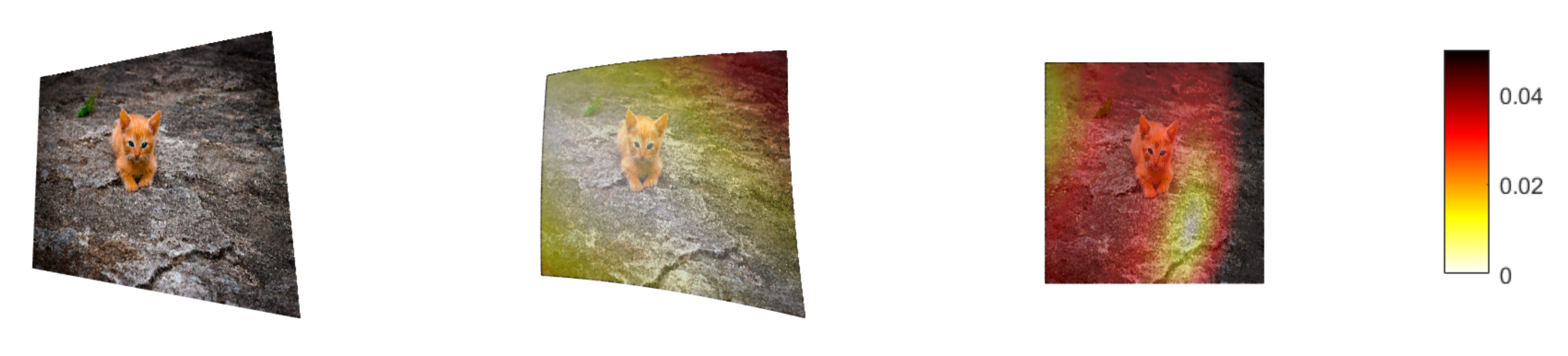}
    \caption{An image (in $\mathbb{P}^2$-domain) to which a rotational homography (left) and the corresponding $\PY$-translation (center) and conventional translation (right) has been applied. The overlay depicts the distortions compared to applying the rotational homography, measured in Euclidean distance of points on the sphere.}
    \label{fig:distortions_image}
\end{figure*}
In Figure~\ref{fig:distortions_image}, we compare the distortions  from using a $\PY$-translation 
and a translation 
in the image plane, respectively, to approximate the action of applying the rotational homography associated with a pitch-yaw $\exp(\alpha)$. The transformations are chosen so that the origin is transformed to the same point as the rotation would do. As can be seen, the $\PY$-group approximates the rotational homography significantly more faithfully than the translation.
\begin{figure}[t]
    \centering
    \includegraphics[width=.5\columnwidth]{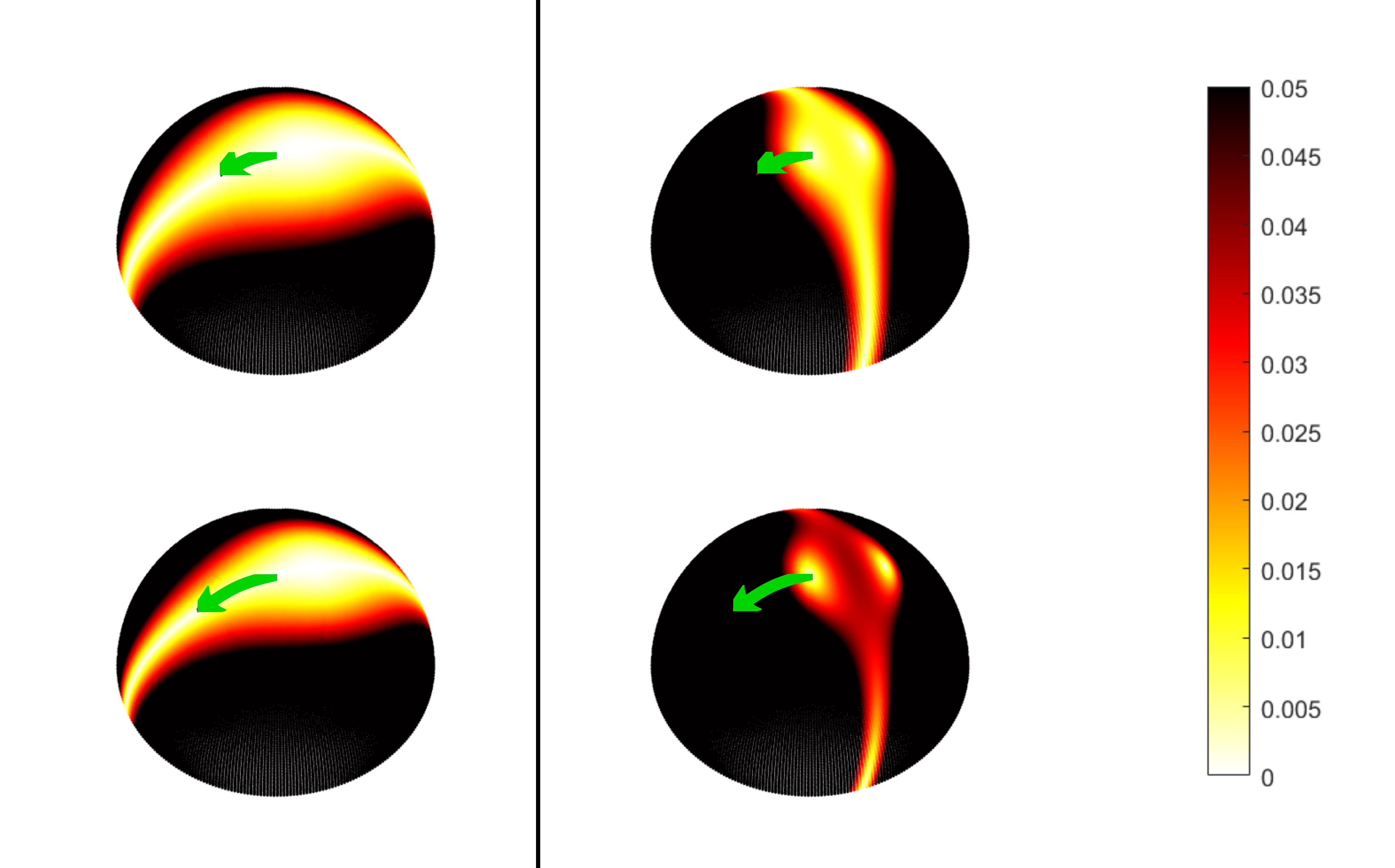}
    \caption{The distortions corresponding to the $\PY$-group (left) compared to the distortion corresponding to the  translation group (right).
    The above plots are for a rotation of $\pi/9$ radians, the below for a rotation of $\pi/6$ radians.
    The respective rotations are illustrated by the green arrows, whose starting points mark $\theta=e_2$.
    \label{fig:distortions}
    }
    
\end{figure}

In Figure \ref{fig:distortions}, we plot the distortions more abstractly, showing the difference in norm of applying an element of the $\PY$-group and a translation in the image plane, respectively, to applying the actual rotation to different points on $\sph^2$. We can observe the following
\begin{itemize}
    \item The distortion of the $\PY$-group has a `sweet area' along the great circle traced out as $e_2$ is rotated by $\exp(\alpha)$.
    \item The distortion of the translation group has a `sweet spot' in a point `opposite' to the direction of the rotation, and a `sweet area' near the equator on a great circle perpendicular to the direction of rotation. 
\end{itemize}
{In Section~\ref{sec:py_appr} of the supplementary material, we formulate as well as prove these statements formally. Note that whereas the $\PY$ action has a `sweet area' along an entire great circle, the sweet area of the translation group lies near the equator, and is hence most often not included in the images. Therefore, these considerations really justify stating that $\PY$ is better at approximating $\SO(3)$ than the translation group is.}

\section{Rotational homography augmentation}\label{sec:aug}
By applying rigidity preserving image transformations, one can generate augmented images and
corresponding
camera poses,
to be incorporated when training a neural network.
In
previous derivations, we assumed a special 3D coordinate system (cf.\ Section~\ref{sec:natural_image_transformations}), which is not valid in practice.
Therefore, we will describe the augmentation procedure in a general setting.

For a given camera with intrinsic calibration, represented by a $3\times 3$ upper triangular matrix $K$, camera pose $R \in \SO(3)$ and camera  $c \in \R^3$, a point $x \in \R^3$ is projected to 
$y \in \mathbb{P}^2$ according to
\[
 \lambda y = K R \begin{bmatrix} \, I_{3} \enspace \mbox{-}c \,  \end{bmatrix} \begin{bmatrix} x \\ 1 \end{bmatrix} ,
\]
where $\lambda$ is a scaling factor 
and $I_{3}$ is the identity matrix, see \cite{hartley-zisserman-book-2004}. If we warp the original image $f$ with the homography
$H=KR_{\mathrm{aug}}K^{-1}$, we will get a new image $f_{\mathrm{aug}}$, where the image coordinates are related by
$y_{\mathrm{aug}} = H y$. Then, the pose for the augmented camera should be $R_{\mathrm{aug}}R$ with unchanged position at $c$, which  can be verified~by checking that the camera equation still holds:  
\[
\lambda y' = \lambda H y = H K R \begin{bmatrix}\,  I \,\mbox{-}c \,  \end{bmatrix} \begin{bmatrix} x \\ 1 \end{bmatrix} 
 = K R_{\mathrm{aug}} R \begin{bmatrix} \, I \, \mbox{-}c \,  \end{bmatrix} \begin{bmatrix} x \\ 1 \end{bmatrix}.  
 \]

\begin{rem}
{
In many computer vision applications, there may be several cameras with different intrinsic calibrations involved. In such cases, it makes sense to augment with the homography $H=K'R_{\mathrm{aug}}K^{-1}$ where $K'$ is another
calibration matrix,
equivalent to performing augmentation with
a general homography $H$.
The
intrinsics
$K$ may vary across training examples, but is always assumed to be known.

In particular, this approach is compliant with uncalibrated pose estimation problems, where both camera intrinsics and extrinsics are unknown at inference and to be estimated simultaneously.
During training one would still require annotations of intrinsic (as well as extrinsic) camera parameters, either separately or recoverable via
RQ-factorization.

Note that the $\PY$-equivariant network approach requires known intrinsics also during inference, and is only suitable for calibrated pose estimation problems.
}
\label{rem:homaug}
\end{rem}

\section{Experiments}\label{sec:exp}
To demonstrate the effectiveness of our proposed sampling scheme and data augmentation,
we carry out experiments on 6D object pose estimation.
The code used in the experiments will be made available.  
We have also carried out experiments on localization, which are deferred to Section~\ref{sec:appendix_localization} of the supplementary material.

\subsection{Datasets}
Experiments are carried out on \LM{} as well as \LMO{}. \LM{} is a standard benchmark for 6D object pose estimation \cite{Hinterstoisser2013}.
The dataset consists of $15$ RGB-D indoor image sequences where objects are laid out on a table with cluttered background.
For each sequence there is a corresponding object of interest put at the center, for which a CAD model is available.
We disregard two of the objects with low quality CAD models, and disregard the depth channel as input, as is the case for \cite{bukschat2020efficientpose} and many works before them.
The \LMO{} dataset was produced by \cite{Brachmann_2014_ECCV} by taking one of the \LM{} sequences and annotating the pose of the surrounding $8$ objects.
These often partially occluded objects make up the objects of interest for this dataset.
For experiments on \LM{}, we conventionally split each sequence into training and test data as e.g.\ \cite{bukschat2020efficientpose,Peng_2019_CVPR,Brachmann_2016_CVPR}, with $\sim 200$ samples for training and $\sim 1000$ samples for test.
For \LMO{}, the very same approach is taken.
Note however that while consistent with \cite{bukschat2020efficientpose}, with which we compare our results, this is not consistent with all other works performing experiments on \LMO{}.
As is conventional, we use  the \add{} metric \cite{Hinterstoisser2013} to evaluate our experiments. 
\subsection{Models}
Our aim of these experiments is to showcase that applying our framework can lead to a performance boost.
As baseline model, we have chosen
\EP{}~\cite{bukschat2020efficientpose},
one of the top-performing models
on the \LM{} dataset,
and for which
the code is
publicly available.

 The size of the \EP{} model is
 governed by a single hyperparameter $\phi$.
Experiments are performed in two settings: (1) pitch-yaw equivariance ($\PY$), and (2) rotational homography data augmentation ($\Hm_{\mathrm{aug}}$).
$\PY$ domain, { as illustrated in Figure~\ref{fig:warp}.}
While \EP{}, and many others, apply geometric data augmentation in the form of image rescale and image rotation, in our second setting we also apply a subsequent pitch-yaw rotational homography during augmentation. 
For further details, see
Section~\ref{sec:impl_ep} of the supplementary material.

Training was
done
in accordance with \cite{bukschat2020efficientpose}, apart from a
few
adjustments.
To speed up training, we used a larger batch-size and adjusted learning rate for the $\phi=0$ model. We made two larger modifications to the
code base:
(1) enabling
reshuffling at every epoch start 
(2) refraining from cherry-picking
the optimal model over all epochs, w.r.t. the \add{} metric on the test set.
For all baseline experiments, models are retrained with these proper modifications, enabling a fair comparison.
For further details, see
Section~\ref{sec:impl_ep} of the supplementary material.

In all experiments, data augmentation is performed as follows.
First, image rescaling is applied with a factor sampled uniformly from $[0.7, 1.3]$.
Furthermore, image plane rotations are carried out around the principal point, with a `roll' angle uniformly sampled from $[-180, 180]$ by default.
This is all consistent with \cite{bukschat2020efficientpose}.
Further, we experiment with more realistic reduced `roll' angles sampled from $[-45, 45]$.
Finally, in the $\Hm_{\mathrm{aug}}$ case, a random homography is applied corresponding to a camera pitch-yaw rotation with an angle sampled uniformly from $[-20, 20]$.

\begin{table}[t]
    \centering
    \scriptsize
\begin{tabular}{|c|c|c||c|c|}
    \hline
    \multirow{2}{*}{Experiment setting}
        & \multicolumn{2}{c||}{$\phi=0$}
        & \multicolumn{2}{c|}{$\phi=3$} \\
     \cline{2-5}
        & Roll 180° & Roll 45°
        & Roll 180° & Roll 45° \\
     \hline
     Baseline
        & \result{77.44} (\result{79.04}) &  \result{76.98}
        & \result{81.94} (\result{83.98}) &  \result{78.92}
        \\
     \hline
     $\PY$
        & \newcorrectcell \result{80.43} & \newcorrectcell \result{80.76} 
        & \result{82.91} &  \result{82.76}
        \\
     \hline
     $\Hm_{\mathrm{aug}}$
        & \result{81.92} & \resultbf{83.58} 
        & \resultbf{85.20} & \result{84.92}
        \\
     \hline
\end{tabular}

    \smallbreak
    \caption{
    Object pose results
    on \LMO{} using the \add{} metric.
    }
    \label{tab:occl_aggregated}
\end{table}

\begin{table}[t]
    \centering
    \scriptsize
\begin{tabular}{|c|c|c|c|c|c|c|c|c|}
\hline

EP:Baseline                    & EP:$\PY$        & EP:$\Hm_{\mathrm{aug}}$
& PVNet\cite{Peng_2019_CVPR}
& CDPN\cite{Li_2019_ICCV}
& BPnP\cite{Chen_2020_CVPR}
& RNNPose\cite{xu2022rnnpose}
& DFPN-6D\cite{Cheng_2021_IEEETransInstrumMeas} \\


\hline
\result{95.56}(\result{97.35}) 
& \newcorrectcell \result{97.43} 
& \result{97.21} 
&  \result{86.27}
& \result{89.86} 
& \multicolumn{1}{c|}{\result{93.27}} 
& \result{97.37} 
& \resultbf{98.06} 
\\

\hline
\end{tabular}
\\

    \smallbreak
    \caption{
    Comparison of object pose results on \LM{} using the \add{} metric, averaged across all objects.
    The EP results (ours) are median results over three runs of the average \add{} across all objects.
    Results reported by \cite{bukschat2020efficientpose} in parentheses.}
    \label{tab:lm_aggregated}
\end{table}

\subsection{Results}
In Table~\ref{tab:occl_aggregated}, results on \LMO{} are presented.
Each reported result is the median score of three training runs.
As expected, there is some
discrepancy between the baseline trained
by us and the original reported results, in particular due to not cherry-picking the best performing epoch (see above).
 In our experiments, the models trained with $\PY$-images perform consistently better than the
baseline and
the
$\Hm$-augmented models perform even better. That augmentation is beneficial is not surprising, considering that the dataset only has $\sim 180$ training images. 

For all further experiments, we used the smaller $\phi=0$ model, as it is much more feasible to train. 
We also choose to moderate the `roll' augmentations to a maximum of 45° when combined with $\Hm_{\mathrm{aug}}$, since this proved superior for the limited capacity $\phi=0$ model on \LMO{}.
For context, we present results on \LM{} with the \add{} metric in Table~\ref{tab:lm_aggregated}, along with a comparison to other
methods.
From these results, we see that by incorporating either our rotational homography augmentation $\Hm_{\mathrm{aug}}$ or the $\PY$-equivariance, the median performance of \EP{} is increased. It should be noted that we do not quite reach the SOTA performance reported in \cite{Cheng_2021_IEEETransInstrumMeas}.
This fact does however not affect the main conclusion of our experiment
-- that our framework can lead to a boost of already very competitive models. Also note that \cite{Cheng_2021_IEEETransInstrumMeas} already apply rotational homography augmentation.

\subsection{Limited data}
\begin{figure}[t]
    \centering
    \includegraphics[width=.8\columnwidth]{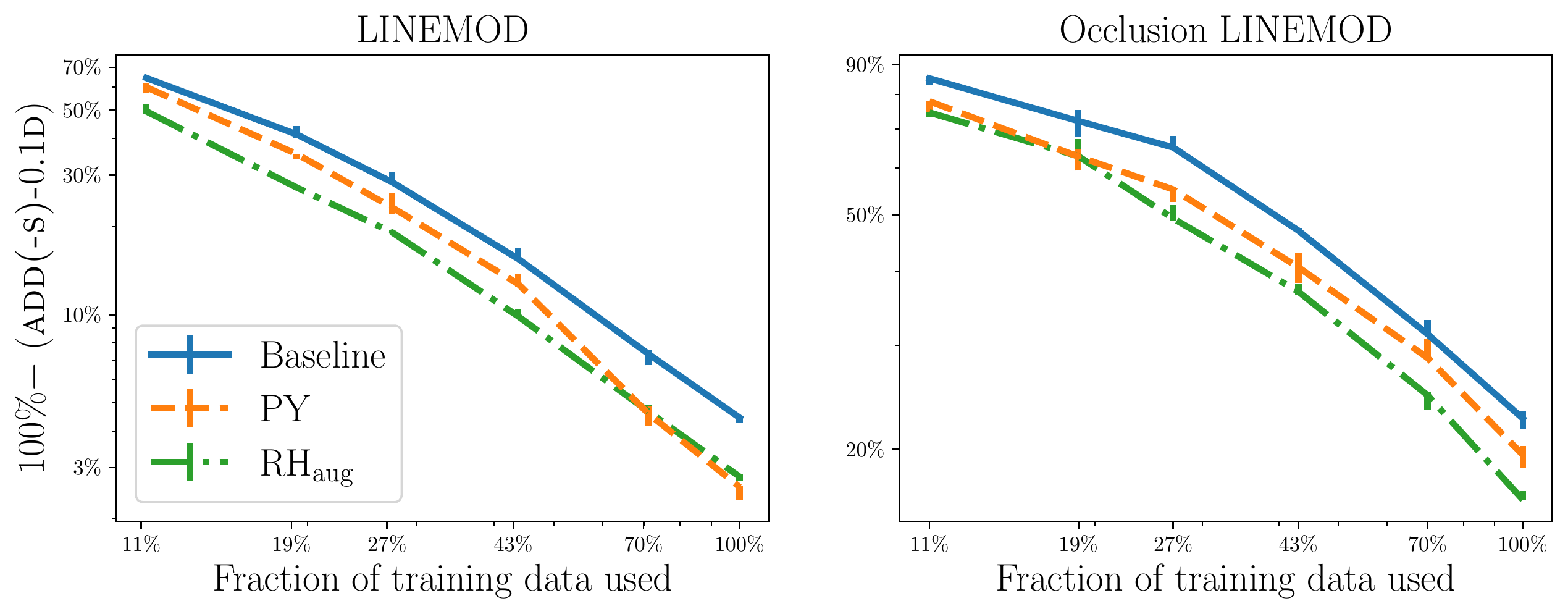}
    \caption{
    Left: Results of 702 limited data experiments on the \LM{} dataset.
    Right: Results of 54 limited data experiments on the \LMO{} dataset.
    In both log-log plots, the line follows the median of three trials, whereas the error bars indicate the best and worst of the trials.
    The \add{} metric is inverted, such that the $y$-axis shows an error function $100 \% - \text{\add{}}$.
    The \add{} is averaged across all objects prior to computing the median over the reruns, as well as the error bars.
    }
    \label{fig:limit_results}
\end{figure}
To further investigate the role of equivariance, we perform
an extensive study where we artificially limit the amount of training data available.
For each sequence in the \LM{} dataset, as well as the \LMO{} dataset,
we create five new training sets by randomly choosing a subset of $n\in\{20, 35, 50, 80, 130\}$ images to train on, respectively (the total number is between $173$ and $189$ depending on the sequence).
Each experiment is repeated three times, to mitigate the issue of statistical fluctuations. We perform 702 experiments on \LM{} and 54 experiments on \LMO{}, training 756 network models in total.
The \LM{} experiments are more since a separate model is trained for each of the 13 objects.
We use the same hyperparameters as in the previous experiments.
In Figure~\ref{fig:limit_results}, we plot the error (i.e. \mbox{$100\%-(\add)$}) against the amount of training data, averaged over the sequences.
Notably, for each of the 6 fractions of training data, we can see that for all 3 reruns, the $\PY$-equivariant model consistently outperforms the 3 retrained instances of the baseline model.
Furthermore, data augmentation often results in even better performance, although in the \LM{} setting the $\PY$-equivariant model is the superior one when training on all or almost all training data.

For the \LM{}-experiments in particular, we note a special effect of the equivariance. The results suggest that the relative performance boost increases with the amount of data for the $\PY$-equivariant nets.  A similar effect of introducing equivariance in a neural network was noticed
in~\cite{batzner2022-e3-nature}
in a completely different context using an entirely different network (calculation of interatomic potentials, which are invariant to rotations of the molecules).
This gives further evidence to the potential of $\PY$-equivariance
for facilitating model generalization.
This effect is not as prominent for  the \LMO{} dataset -- we speculate that for this more delicate learning task the dataset is too small for the model to enter the region where the equivariance effect kicks in. Note that in the \LM{} plot, the $\PY$-model error only falls steeper in the region with more data.

In Section~\ref{sec:results_per_object} of the supplementary material further results are presented, for each of the objects separately.

\section{Conclusions}
We present
a framework for handling perspective effects in images taken by a pinhole camera. More specifically, we provide a mathematical argument for why translation equivariance is not the ideal objective and that it is more natural to either design networks that are
equivariant to camera rotations
simulate camera rotations when performing data augmentation.
As an alternative to true equivariance with respect to camera rotation, we also present and analyse a group structure ($\PY$) on the set of rotations around axes in the image plane. Our framework is generally applicable and can easily be integrated into existing CNN architectures.
We have tested our framework on 6D object pose estimation using a method relying on a CNN backbone and
our
experiments indicate that increased rotation equivariance can result in a substantial performance boost.

\paragraph{Acknowledgement:} The authors acknowledge support from Chalmers AI Research (CHAIR). This work was partially supported by the Wallenberg AI, Autonomous Systems and Software Program (WASP) funded by the Knut and Alice Wallenberg Foundation, and also the Swedish Foundation for Strategic Research
(Semantic Mapping and Visual Navigation for Smart Robots)
 The computations were enabled by resources provided by the Swedish National Infrastructure for Computing (SNIC) at Chalmers Centre for Computational Science and Engineering (C3SE) partially funded by the Swedish Research Council through grant agreement no. 2018-05973.

{\small
\bibliographystyle{splncs04}
\bibliography{newbib.bib}
}

\newpage
\appendix

\section{Implementation details -- EfficientPose}\label{sec:impl_details}
\label{sec:impl_ep}
This section contains implementation details that were left out of the main paper due to space limitations.

\subsection{Pitch-yaw equivariance}\label{sec:impl_details_py}

Here, we describe in detail how to modify the images to achieve a $\PY$-equivariant architecture. In short, the idea is to
transform
the images into the $\PY$ domain, and subsequently apply the CNN.
    As outlined in Proposition~\ref{prop:warp} in Section~\ref{sec:proofs_warp}, the operation in question can be described by transforming the (calibrated) image coordinates according to $\phi_{\mathrm{pol}}(\tan(r),\varphi ) \mapsto \phi_{\mathrm{pol}}(r,\varphi- \tfrac{\pi}{2})$.
    In practice, we disregard the resulting image plane rotation by $-\tfrac{\pi}{2}$, resulting in a purely non-linear rescaling of the radius according to $\tan(r) \mapsto r$.
    Finally, we transform our image into pixel coordinates by applying an affine transformation in each of the coordinates. These are chosen such that
    the transformed
    image has the same size as the original image, while including as few pixels  where the image is undefined as possible. Note that in our special setting, where all images in the database are taken by the same camera,
    this amounts to sampling all $\PY$-images on a common 'exhausting' grid in $\PY$-space. For more diverse datasets, more care needs to be taken to achieve this.

    The  pixels in which the the resulting images is undefined are simply set to zero. The object segmentation masks are
    transformed
    along with the images, effectively resulting in a transformation of the resulting bounding boxes as well. A graphical description of the procedure is given by Figure~\ref{fig:warp} in the main article. In order to increase readability, we will refer to this procedure as \emph{warping} the image.

    In addition to transforming the images, we also make a couple of adjustments to the prediction targets, with the purpose of making the learning task
    suitable for
    the $\PY$-equivariant setting.
    The labels are rigid motions $(R,t) \in \SO(3) \times \R^3$, which describe the pose of the objects relative to the camera coordinate system.
    Each translation target $t\in \R^3$ (the position of an object in the camera frame) is transformed to $(\alpha,s)$, where $\alpha \in\PY$ is the warped version of $\pi(t)$, and $s=\abs{t}\in \R$ is the camera-to-object distance.
    As compared to $t_2$, we believe the distance between object and camera to be a more fitting target for the
    approximate spherical camera model
    in our
    approach. To be precise, while for a standard pinhole camera the object appearance in the image (in particular, its size) is roughly determined by $t_2$, this is not the case for a spherical camera.
    In this case, the camera-to-object distance is instead the corresponding determining factor.

    Note that the new representation shares some properties with \cite{bukschat2020efficientpose}: The authors of that article also decouple the prediction of  $t_0, t_1$ from the depth $t_2$. Rather than predicting the coordinates $t_0$  $t_1$ directly, an offset is predicted from the center of an anchor box to the projection of $t$ in the image.
    Our adjustments for the $\PY$ case are  consistent with this, with the difference that
    both anchor box centers and $t$ projections are expressed in the $\PY$-domain, and consequently the offsets as well.

    Similarly, we argue  that in the $\PY$ case, the object orientation $R\in \SO(3)$ should not be expressed as a rotation relative to the camera frame.
    Instead, we define a pixel-dependent frame of reference  $S$ for rotation estimation.
    Such a coordinate frame is given by taking the camera frame, and rotating it such that the
    $x_2$ axis,
    previously aligned with the optical axis, now aligns with the viewing ray corresponding to the pixel.
    Such a rotation is of course not uniquely determined; we use the one of the form $\exp(\alpha)$, $\alpha \in \PY$, with $\abs{\alpha}$ minimal.
    The adjustment of the rotation target is motivated by the fact that an object's appearance in $\PY$ is roughly invariant to rotations around the camera (only translated), and should only be inferred relative to such a varying frame.
    There is further evidence in the literature supporting this choice. In \cite{kundu_3d-rcnn_2018} the same rotation targets as we propose are used with success for 3D object reconstruction.

\subsection{Image rescale augmentation}
 \EP{} proposes a delicate way to scale-augment $6D$-pose data.  When scaling the image by a factor $f$, the authors propose to scale
 the target depth coordinates $t_2$
 by the factor $\tfrac{1}{f}$ (effectively modeling an $x_2$-translation), as an object farther away ($f<1$) will appear smaller in the image (and vice versa).
 As the authors of \cite{bukschat2020efficientpose} point out, this is an approximation, which will introduce an error unless $t_2$ accurately describes a constant depth for the whole visible object surface.
 Occlusion effects can also arise, especially for objects far from the center of the image.

Likewise, we adopt an approximate scaling augmentation procedure to our setting as follows. Recall that the pose labels are of the form  $(R,\alpha, s) \in \SO(3) \times \PY \times \R$. When we scale a point $(x,y) \in \mathbb{P}^2$ negatively, it will move towards the center of the image. To mimic this behaviour, we choose to transform $(\alpha,s) \in \PY \times \R$ to $(f\alpha, s/f) \in \PY \times \R$. In this manner, moving the object away from the camera ($f<1$) will result in a shift of the object towards the image center. The shift in $\PY$-coordinates is hence equal to $\alpha(f-1)$.

This means, approximately, that we rotate the object according to $\exp(\alpha(f-1))$. We therefore adjust the  rotation $R$ relative to the camera coordinate system according to $R \mapsto \exp(\alpha (1-f))R$.

\subsection{Rotational homography data augmentation}\label{sec:impl_details_homo_aug}
Let us describe the details of the augmentations as described in Section \ref{sec:aug}. Note that three different augmentations are performed: Scalings, rotations in the image plane (referred to as \emph{rolls}) and homographies corresponding to pitch-yaw rotations $\exp(\alpha)$ for $\alpha \in \PY$. The scalings are applied as described in the previous section, with the $f$ parameter uniformly sampled between $0.7$ and $1.3$, i.e. just as in \cite{bukschat2020efficientpose}. The rolls are sampled uniformly in either $[-45\text{°}, 45\text{°}]$ or $[-180\text{°}, 180\text{°}]$. As for the tilt, we  sample the radius of the $\alpha$-parameter uniformly in $[0, 20\text{°}]$, and the direction in the unit plane uniformly over the unit circle. 

All three augmentations are combined into one single homography $H$, in order to reduce interpolation errors and truncation effects.
 \subsection{Non-geometric augmentations} In addition to the geometric augmentations described above, the same color-space augmentations as in \cite{bukschat2020efficientpose} are applied. These are essentially the ones given by the \texttt{RandAugment}-function of the imgaug-package \cite{imgaug_github}, however only including the augmentations in the color space, i.e. excluding crops, rotations, shears etc.
 
 \subsection{Further changes to the code}
 In difference to \cite{bukschat2020efficientpose}, we reshuffle the data between epochs, and only record the performance of the last model, and thus do not cherry-pick the best performing epoch.
 These changes are applied for all experiments, including the baseline, to facilitate a fair comparison.
 
 \subsection{Hyperparameters}
 We use essentially the same hyperparameters as \cite{bukschat2020efficientpose}, with a notable exception for the $\phi=0$ experiments.
 These used a learning rate of $3 \cdot 10^{-4}$, instead of $10^{-4}$, and a batch size of $6$, instead of $1$.
 This change was only made to speed up the experiments.
 Furthermore, we explored a moderation of the range of `roll' augmentation from a maximum of 180 degrees to 45 degrees, and applied this moderation for all $\Hm_\mathrm{aug}$ experiments with the small $\phi=0$ model, unless mentioned otherwise. We have made no further hyperparameter modifications as compared to the original implementation.

 We initialize with pretrained parameters of an \ED{} model \cite{Tan_2020_CVPR} trained on COCO \cite{coco}, and train with the Adam optimizer, using gradient norm clipping with $10^{-3}$ threshold, and a learning rate decay schedule, all consistent with \cite{bukschat2020efficientpose}. The number of training epochs depends on the fraction of data used -- we scale up the number of epochs for low data fractions in order to have enough training iterations for the loss to converge. For the full data we run 5k epochs, as did \cite{bukschat2020efficientpose}. For the 20, 35 and 80 images cases we run 10k epochs and for the 130 image case we run 7k epochs. We cap the epoch number at 10k as it seems to suffice for convergence in all cases
 Validation is carried out every 100 epochs for the $\phi=0$ experiments, and every 50 epochs for $\phi=3$.

\subsection{Hardware and runtime}\label{sec:ep_hardware}
We use two computing setups.
Initial \LMO{} experiments were carried out with Tensorflow 1.15 in a Docker container running Ubuntu 18.04, on a workstation with host operating system Manjaro. The hardware has 64 GB RAM, a 6-core Intel Core i7-8700K CPU, and an Nvidia GTX 1080 Ti GPU with 11GB memory.
The \LM{} experiments use the same Docker container\footnote{However, the Docker container is here converted to a Singularity container due to system constraints.}, but these experiments involve many training cycles since a separate model is trained for each object, and therefore computations are carried out on a compute cluster.
Further runs of the \LMO{} experiments are also carried out on the compute cluster in the same way.
Each compute node is running CentOS 7 and is equipped with 72GB RAM, four CPU cores of type Intel(R) Xeon(R) Gold 6226R CPU @ 2.90GH, and an Nvidia Tesla T4 GPU with 16GB memory
Training takes around 30 hours for each $\phi=0$ model on the compute cluster, and around 24 hours on the local workstation.
Each $\phi=3$ takes around 1 week to train.

\section{A group action of $\PY$ on $\N \times \mathbb{S}^2$} \label{sec:groupaction}

Here, we describe how an action of $\PY$ on $\N \times \mathbb{S}^2$ can be constructed. Recall that an action of a group $G$ on a set $K$ is a map $\chi: G \times K \to K, (g,k) \mapsto \chi_g(k)$ with the property that $\chi_g(\chi_h(k))=\chi_{gh}k$. 

\begin{figure}
    \centering
    \includegraphics[width=.5\textwidth]{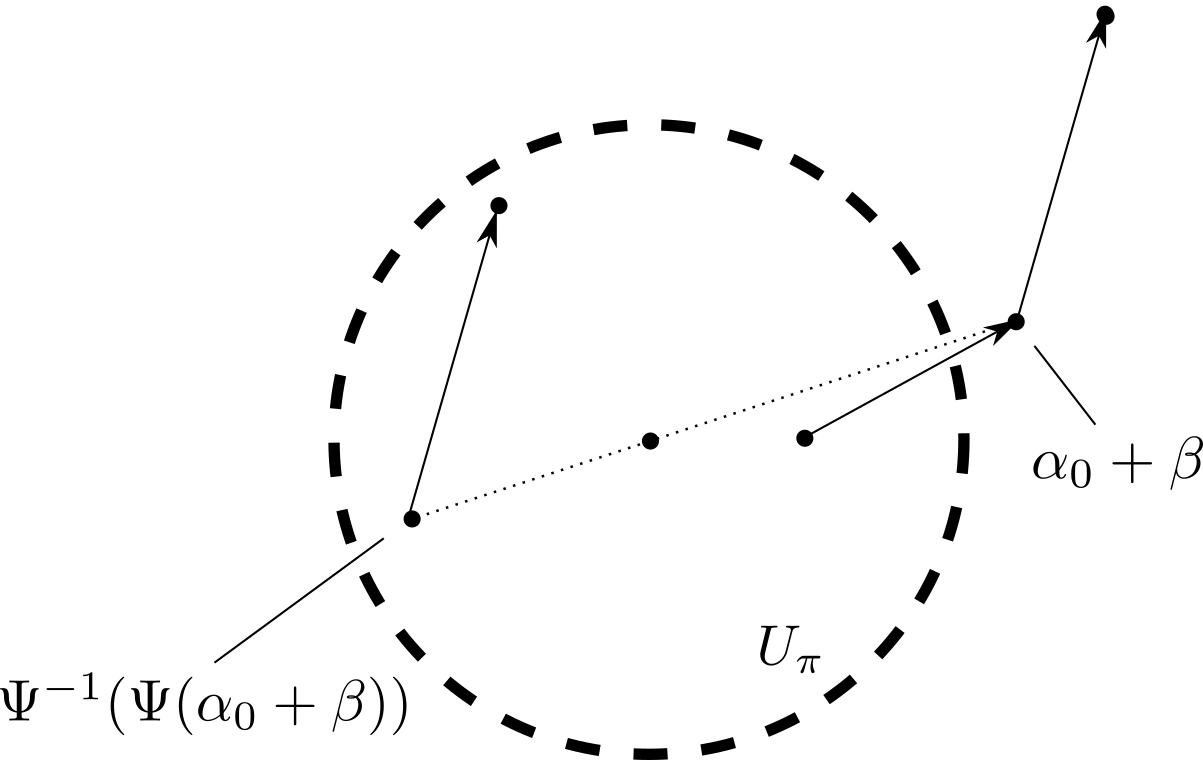}
    \caption{$\Psi^{-1}(\Psi(\alpha_0+\beta)) + \gamma \neq \alpha_0 +\beta + \gamma$ \label{fig:non_group}}
\end{figure}

The action should fulfill satisfy $\chi_\alpha(e_2) = \exp(\alpha)e_2$, at least for small $\alpha$, in order to resemble the $SO(3)$-action of the pitch-yaw $\exp(\alpha)$ as good as possible close to $e_2$. A natural ansatz for the group action would be, defining the map $\Psi: \PY \to \sph^2$ through $\Psi(\alpha)=\exp(\alpha)e_2$, the following formula:
\begin{align*}
    \chi_\beta(\theta) = \Psi(\beta + \Psi^{-1}(\theta)),
\end{align*}
where the inverse is the inverse of the restriction of $\Psi$ on $U_\pi$. 
However, this is \emph{not} a group action. To see this, note that, excluding the circles of radius $n \pi, n \in \N$ around the origin, the map $\Psi^{-1} \circ \Psi$ maps $\alpha$ to the unique element on the line
\begin{align*}
    \ell_\alpha = \{ \beta \, \vert \, \beta \text { collinear with $\alpha$ }, \abs{\beta}-\abs{\alpha}  \in 2\pi \Z \}
\end{align*}
which lies inside $U_\pi$. Hence, for $\theta= \Psi(\alpha_0)$ with $\alpha_0 \in U_\pi$, $\Psi^{-1}(\chi_\beta(\theta)) = \Psi^{-1}(\Psi(\alpha_0+\beta)) \neq \alpha_0 +\beta$ for many $\beta$. In fact, for such $\alpha_0$ and $\beta$, we will even for most values for $\gamma$ have $\gamma \neq \alpha_0 +\beta + \gamma$ (see Figure~\ref{fig:non_group}). Hence, 
\begin{align*}
    \chi_{\gamma}(\chi_{\beta}(\alpha_0)) &= \Psi(\gamma + \Psi^{-1}(\chi_\beta(\theta)) \\
    &= \Psi(\gamma +  \Psi^{-1}(\Psi(\alpha_0+\beta))) \\
    &\neq \Psi^{-1}(\gamma + \alpha_0 +\beta) = \chi_{\gamma+ \beta}(\theta)
\end{align*}
This explains why need to define a map on $\PY$ with better bijectivity properties. This is possible if we extend the range to $\N \times \mathbb{S}^2$.

\begin{prop} \label{prop:PYparametrization}
     For $[\alpha] \in \PY$, we consider the curve $\gamma: [0,1]\to \sph^2, t \mapsto \exp(t\alpha)e_2$. Letting $n$ be the number of times $\gamma$ crosses the set $\{\pm e_2\}$, we define a map $\Phi: \PY \to \N \times \sph^2$  through
     \begin{align*}
         \Phi(\alpha):=(n,\gamma(1)).
     \end{align*}
     Then, $\Phi$ is injective on $\PY$, excluding circles of radius of multiples of  $\pi$ around $0$.
\end{prop}

To prove this lemma, let us first show that $\gamma(t)=\exp(\alpha t)$ defines a geodesic.
\begin{lemma} \label{lem:geodesics}
    For $\alpha \in \PY$, $\gamma: \R \to \sph^2, \, t \mapsto \exp(t\alpha)e_2$ is a geodesic with speed $\abs{\alpha}$ and $\gamma'(0) =\alpha e_2$.
\end{lemma}
\begin{proof}
   Due to the properties of the matrix exponential map, the first two extrinsic derivatives of $\gamma$ are given by
   \begin{align*}
       \gamma'(t) = \alpha \exp(\alpha t) e_2 , \quad \gamma''(t) = \alpha^2 \exp(\alpha t) e_2.
   \end{align*}
   This already proves that $\gamma'(0)=\alpha e_2$. To show that $\gamma$ is a geodesic, we now only need to argue that $\gamma''(t)$ is perpendicular to the tangent plane at each point $\gamma(t)$. For the sphere, this simply means that $\gamma''(t)$ should be parallel with $\gamma(t)$. However, since $\alpha$ commutes with $\exp(\alpha)$, we get
   \begin{align*}
       \gamma''(t) = \alpha^2 \exp(\alpha t) e_2 = \exp(\alpha t) \alpha^2 e_2 = \abs{\alpha}^2 \exp(\alpha t) e_2 = -\abs{\alpha}^2 \gamma(t),
   \end{align*}
   where the penultimate step is the result of a simple computation. The proof is finished.
\end{proof}

Proving Lemma \ref{prop:PYparametrization} is now easy.
\begin{proof}[Proof of Prop. \ref{prop:PYparametrization}]
    Due to Lemma \ref{lem:geodesics}, we know that for each $\alpha$, $\gamma$ is a geodesic with speed $\abs{\alpha}$. Therefore, it traverses along a great circle which goes through $e_2$. This great circle is uniquely determined by $\gamma(1)$, unless $\abs{\alpha}$ is a multiple of $\pi$, which we excluded. Thus, if two elements have the same image under $\Phi$, the corresponding curves must have traversed along the same great circles. The only way they could have ended up in the same endpoint $\gamma(1)$ is that one curve has traversed the geodesic a multiple of a half revolution of the entire sphere more times. However, since they have crossed $\{\pm e_2\}$ the same number of times, the curves must however be exactly equal,  (which for instance can be seen by comparing derivatives in $t=0$), and therefore also the endpoints.
\end{proof}

Using $\Phi$, the action may now (excluding the south poles $(n,-e_2), n \in \N$) be defined as follows
\begin{align*}
    \chi_\beta(\theta) = \Phi(\beta + \Phi^{-1}(\theta)).
\end{align*}
Since $\Phi$ is injective, we then have $\Phi^{-1}\circ \Phi(\alpha_0+\beta)=\alpha_0 +\beta$ for all $\alpha$ with $\abs{\alpha} \neq \pi \N$, which prevents the problems described above from occuring.

\section{Approximation property of \texorpdfstring{$\PY$}{PY}} \label{sec:py_appr}

In this section we provide a formal analysis of the approximation property of the $\PY$-group. Fix an $\alpha \in \PY$. We will compare three maps
\begin{align*}
 \psi_\alpha &: \sph^2 \to \sph^2, \theta \mapsto \exp(\alpha)\theta, \\
        \varphi_\alpha &:  \sph^2 \to \sph^2, \theta \mapsto \mathrm{pr}_{\sph^2} \Phi\left(\alpha +  \Phi^{-1}(0,\theta)\right), \\
        \chi_\alpha &: \sph^2 \to \sph^2, \theta \mapsto \pi^{-1}(\pi(\theta) +t),
\end{align*}
where $t$ is defined via $\pi(\exp(\alpha)e_2)$, and $\mathrm{pr}_{\sph^2}$ projects $(n,\theta)\in \N\times \sph^2$ to $\theta \in\sph^2$. Note that the inverse of $\pi$ here refers to its restriction to the upper half of $\sph^2$, in formulas
\begin{align*}
    \pi^{-1}: \mathbb{P}^2 \to \sph^2, v \mapsto \tfrac{1}{\sqrt{1+\abs{v}^2}} \begin{bmatrix} v \\ 0 \end{bmatrix}.
\end{align*} 
The three maps $\psi_\alpha, \varphi_\alpha, \chi_\alpha$ correspond to actually applying the pitch-yaw rotation ($\psi_\alpha$), applying the pitch-yaw as a $\PY$-translation ($\varphi_\alpha$), and applying the translation in the image plane
such that it is translated
as it would have been by the rotation and subsequent projection onto $\mathbb{P}^2$ ($\chi_\alpha$). 

Already comparing the plots of the functions $\abs{\varphi_\alpha-\psi_\alpha}$ and $\abs{\chi_\alpha -\psi_\alpha}$ in Figure~\ref{fig:distortions} in the main article, we see that the $\PY$-translation $\varphi_\alpha$ is much better at approximating the rotation $\psi_\alpha$ than the translation $\chi_\alpha$. In the following, we provide a theoretical analysis, which will prove additional insight.

We begin by comparing $\psi_\alpha$ and $\varphi_\alpha$, i.e. the action of $\exp(\alpha)$ as an element in $\SO(3)$ and of $\alpha$ as a  $\PY$-translation.
\begin{prop} \label{prop:PYvssO}
    We have
    \begin{equation*}
        \abs{\varphi_\alpha(\exp(\beta)e_2)-\psi_\alpha(\exp(\beta)e_2)} \leq  \abs{\tfrac{1}{6}[\alpha,\beta](\alpha + 2\beta) e_2} + \mathcal{O}( \abs{\alpha}^3) + \calO( \abs{\beta}^3).
    \end{equation*}
    Here, $[\alpha,\beta]$ denotes the commutator $(\alpha \beta - \beta \alpha)\in\R^{3,3}$.
\end{prop}

\begin{proof}
    We have $\varphi_\alpha (\exp(\beta))= \Phi(\alpha + \beta) =\exp(\alpha+\beta)e_2$. Thus, what we need to estimate is
    \begin{align*}
       \exp(\alpha+\beta)e_2 -  \exp(\alpha)\exp(\beta) e_2.
    \end{align*}
    We have the following Taylor-expansion of $\exp$ around $0$
    \begin{align*}
        \exp(\beta)= \id + \beta+ \tfrac{1}{2} \beta^2 + \calO(\abs{\beta}^3).
    \end{align*}
    We can also Taylor-expand $\exp$ around $\alpha$ to obtain
    \begin{align*}
        \exp(\alpha+\beta)= \exp(\alpha) + \exp'(\alpha)\beta + \tfrac{1}{2}\exp''(\alpha)(\beta,\beta) + \calO(\abs{\beta}^3).
    \end{align*}
    We now need to calculate these derivatives. For this, we use the following formula (see eq. 3 in \cite{NAJFELD1995321})
    \begin{align*}
        &\exp^{(k)}(t\alpha)(t\beta)^k \\&= k\cdot \exp(t\alpha) \int_0^t \exp(-s\alpha) \beta\exp^{(k-1)}(s\alpha)(s\beta)^{k-1} \dd s, \quad \exp^{(0)}(t\alpha) \\
        &= \exp(t\alpha).
    \end{align*}
    Let us begin by investigating the first derivative. Since
    \begin{align}
        &\exp(-s\alpha)\beta\exp(s\alpha)  \\ &=\left(\id -s\alpha  +\tfrac{s^2}{2}\alpha^2+ \calO(s^3\abs{\alpha}^3)\right) \beta\left(\id +s\alpha  +\tfrac{s^2}{2}\alpha^2+\calO(s^3\abs{\alpha}^3)\right) \label{eq:commutation}\\
        &= \beta -s[\alpha,\beta]+ \tfrac{s^2}{2} [\alpha,[\alpha,\beta]]+ \calO(\abs{\alpha}^3) \nonumber
    \end{align}
    we get by carrying out the integration
    \begin{align} \label{eq:firstderivative}
        \exp'(t\alpha)t\beta = \exp(t\alpha)\left(t\beta - \tfrac{t^2}{2}[\alpha,\beta] + \tfrac{t^3}{6}[\alpha,[\alpha,\beta]] + \calO(\abs{\alpha}^3)\right).
    \end{align}
    We may now use this to estimate the second derivative
   
    \begin{align} 
        &\tfrac{1}{2}\exp''(\alpha)(\beta,\beta) \\
        &= \exp(\alpha)\int_0^1\exp(-s\alpha) \beta \exp'(s\alpha)s\beta \dd s  \nonumber\\
        &= \exp(\alpha)\int_0^1\exp(-s\alpha) \beta \exp(s\alpha)
        \left(s\beta - \tfrac{1}{2}s^2[\alpha,\beta] + \tfrac{s^3}{6} [\alpha,[\alpha,\beta]] \right) \dd s \nonumber \\
        & \qquad \qquad  \qquad + \calO(\abs{\alpha}^3) \nonumber \\
        &= \exp(\alpha)\int_0^1 s \beta^2 - \tfrac{s^2}{2}(\beta[\alpha,\beta]+2[\alpha,\beta]\beta  ) \nonumber \\
        & \qquad \qquad \qquad  + \tfrac{s^3}{6}(\beta[\alpha,[\alpha,\beta]]+3[\alpha,[\alpha,\beta]]\beta  + 3[\alpha,\beta]^2) +  \calO(\abs{\alpha}^3) \dd s \nonumber\\
        &=  \exp(\alpha) \bigg(\tfrac{1}{2}\beta^2 - \tfrac{1}{6}(\beta[\alpha,\beta]+2[\alpha,\beta]\beta ) \nonumber \\
        &\qquad \qquad \qquad + \tfrac{1}{24}(\beta[\alpha,[\alpha,\beta]]+3[\alpha,[\alpha,\beta]]\beta  + 3[\alpha,\beta]^2) +\calO(\abs{\alpha}^3)\bigg) \label{eq:secondderivative}
    \end{align}
    where we again used \eqref{eq:commutation} in the penultimate step. Using \eqref{eq:firstderivative} and \eqref{eq:secondderivative}, we get that the difference $\exp(\alpha+\beta)-\exp(\alpha)\exp(\beta)$ equals 
    \begin{align*}
         \exp(\alpha)\bigg( &  - \tfrac{1}{2}[\alpha,\beta]+ \tfrac{1}{6}[\alpha,[\alpha,\beta]] -\tfrac{1}{6}\left(\beta[\alpha,\beta] + 2[\alpha,\beta]\beta \right) \\
         & + \tfrac{1}{24}\left(\beta[\alpha,[\alpha,\beta]]+3[\alpha,[\alpha,\beta]]\beta  + 3[\alpha,\beta]^2\right) + \calO(\abs{\alpha}^3) + \mathcal{O}(\abs{\beta}^3)\bigg)  
    \end{align*}
    
    We now need to remember that we are not looking for an estimate of the matrices, but rather their action on $e_2$. The commutator $[\alpha,\beta]$ must be a multiple of $C_2$, the matrix that represents the linear map $v \mapsto e_2 \times v$. As such, 
    \begin{align*}
        [\alpha,\beta]e_2 = 0.
    \end{align*}
    for all $\alpha, \beta \in \PY$. Using this, the above simplifies to
    \begin{align*}
        \exp(\alpha)&\left(-\tfrac{1}{6}[\alpha,\beta](\alpha e_2 + 2\beta e_2- \tfrac{3}{4} \alpha \beta e_2) + \tfrac{1}{24}(-\beta[\alpha,\beta]\alpha e_2 +3\alpha[\alpha,\beta ]\beta e_2 \right)  \\
        &+ \calO(\abs{\alpha}^3) + \calO(\abs{\beta}^3).
    \end{align*}   
    Now, it is a straightforward calculation to show  that $\alpha \beta e_2 \in \mathrm{span}(e_2)$. As such, we must have $[\alpha,\beta]\alpha\beta e_2=0$. This, together with $[\alpha,\beta]e_2=0$, yields $\alpha\beta \alpha\beta e_2 = \beta\alpha\beta \alpha e_2$. Using this, and the fact that $\alpha^2 \beta^2 e_2=\beta^2\alpha^2e_2 = \abs{\alpha}^2\abs{\beta}^2 e_2$, finally shows that
    \begin{align*}
        \beta[\alpha,\beta ]\alpha e_2 = \beta \alpha \beta \alpha e_2  - \beta^2 \alpha^2 e_2 = - (\alpha^2\beta^2 - \beta \alpha \beta \alpha) e_2 = - \alpha [\alpha, \beta]\beta e_2.
    \end{align*} 
    The error thus equals
    \begin{align}
       -\tfrac{1}{6} \exp(\alpha)( [\alpha,\beta](\alpha e_2 + 2\beta e_2 )+ \beta[\alpha,\beta] \alpha e_2) + \mathcal{O}(\abs{\alpha}^3) + \mathcal{O}(\abs{\beta}^3). \label{eq:errorpre}
    \end{align}
    Now notice that
    \begin{align*}
        \exp(\beta)[\alpha,\beta](\alpha + 2\beta) & = (\id + \beta)[\alpha,\beta] (\alpha + 2\beta) + \mathcal{O}(\abs{\beta}^3)  \\
        &= [\alpha,\beta](\alpha + 2\beta) + \beta [\alpha,\beta]\alpha + \mathcal{O}(\abs{\beta}^3).
    \end{align*}
    Hence, \eqref{eq:errorpre} simplifies to 
    \begin{align*}
        - \tfrac{1}{6}\exp(\alpha) \left( \exp(\beta) [\alpha,\beta](\alpha + 2\beta) + \mathcal{O}(\abs{\alpha}^3) + \mathcal{O}(\abs{\beta}^3)\right)
    \end{align*}
    which, together with the fact that $\exp(\alpha)$ and $\exp(\beta)$ are orthogonal matrices, yields the claim.
    \end{proof}

Notice the $[\alpha,\beta]$-term appearing in the error. This is identically equal to zero for $\beta$ parallel with $\alpha$. This shows, as claimed in the main text,that close to the great circle $t \mapsto \exp(\alpha t)$, the distortion of $\PY$-translating with $\alpha$ compared with performing the actual rotation $\exp(\alpha)$ is especially small. In fact, it is not hard to show that on that great circle, we even have $\psi_\alpha=\varphi_\alpha$. 

 \begin{prop}
     On the great circle $t \mapsto \exp(t\alpha)$, $\varphi_\alpha$ and $\psi_\alpha$, as defined in Proposition \ref{prop:PYvssO}, are equal.
 \end{prop}
\begin{proof}
    We need to compare $\varphi_\alpha(\exp(t\alpha ))=\exp(\alpha +t\alpha)e_2$ and $\psi_\alpha(\exp(\alpha t))=\exp(\alpha)\exp(t\alpha)e_2$. It is however well-known that $\exp(\alpha+ \beta) =\exp(\alpha)\exp(\beta)$ if $\alpha$ and $\beta$ commute, which $\alpha$ and $t\alpha$ surely do.
    This in particular implies
    $\varphi_\alpha(\exp(\alpha t)) = \exp(\alpha +t\alpha)e_2 = \exp(\alpha)\exp(t\alpha)e_2 = \psi_\alpha(\exp(\alpha t))$,
    i.e., the statement. 
\end{proof}

We move on to the comparison of $\psi_\alpha$ and $\chi_\alpha$. We again begin with a general expansion.

 \begin{prop} \label{prop:translvsSO}
      Let $\pi$ be the projection $\sph^2_+ \to \mathbb{P}^2$, $\alpha \in \PY$ and $t$ be defined via  
      $t=\pi(\exp(\alpha)e_2)$
      \begin{align*}
          \chi_\alpha : \sph^2 \to \sph^2, \theta \mapsto \pi^{-1}(\pi(\theta) +t)
      \end{align*}
      Then
      \begin{align*}
          &\chi_{\alpha}(\theta)-\psi_\alpha(\theta) \\
          &=- \langle t, \theta\rangle (t+ \widehat{\theta})   -\tfrac{1}{2}\langle{\theta,\alpha}\rangle \alpha + (\theta_2-1) \tfrac{\abs{t}^2}{2}e_2 + \calO(\abs{t}^3) +  \calO(d_\sph(\theta,e_2)^3) 
      \end{align*}
     where $\langle \cdot, \cdot \rangle$ denotes scalar product, $\psi_\alpha$ is as in Proposition \ref{prop:PYvssO}, $\widehat{\theta}$ the projection of $\theta$ onto $\mathrm{span}(e_0,e_1)$ and we identified the two-dimensional vectors $\alpha$ and $t$ as three-dimensional vectors in $\mathrm{span}(e_0,e_1)$.

    \end{prop}
    
        \begin{proof}[Proof of Proposition \ref{prop:translvsSO}] Let us begin by relating $\alpha$ to $t$. By Taylor-expanding $\pi^{-1}$ in $0$ (viewing it as a function $\mathbb{P}^2 \to \R^3)$, we obtain
       \begin{align*}
           \pi^{-1}(t) = \pi^{-1}(0) + \begin{bmatrix} t  \\ 0 \end{bmatrix} - \tfrac{\abs{t}^2}{2} e_2  +\mathcal{O}(\abs{t}^3) = e_2  + \begin{bmatrix} t  \\ 0 \end{bmatrix} - \tfrac{\abs{t}^2}{2} e_2  +\mathcal{O}(\abs{t}^3) .
       \end{align*}
       Comparing this to 
       \begin{align*}
           \exp(\alpha)e_2 = e_2 + \alpha e_2 + \tfrac{1}{2}\alpha^2 e_2+ \mathcal{O}(\abs{\alpha}^3) =  e_2 + \alpha e_2 - \tfrac{\abs{\alpha}^2}{2} e_2+ \mathcal{O}(\abs{\alpha}^3)
       \end{align*}
       we see that $\exp(\alpha)e_2 = \pi^{-1}(t)$ implies that 
       \begin{align} 
        t= \alpha e_2 + \calO(\abs{\alpha}^3) = \begin{bmatrix} \alpha_1 \\ -\alpha_0 \\ 0 \end{bmatrix}+  \calO(\abs{\alpha}^3). \label{eq:tvsalpha}
        \end{align}
       
       Now, Taylor-expanding $\pi^{-1}$ in $t$ and disregarding all terms which have a least order $3$ in $t$ or $v$, we obtain 
       \begin{align*}
           &\pi^{-1}(t + v) \\
           &= \pi^{-1}(t)  - \left(\langle t, v \rangle + \tfrac{\abs{v}^2}{2}\right) \begin{bmatrix}
           t \\ 1
           \end{bmatrix}\\
           &\quad + \left(1- \tfrac{\abs{t}^2}{2}-  \langle{t,v} \rangle -\tfrac{\abs{v}^2}{2} \right)\begin{bmatrix} v\\ 0 \end{bmatrix}  + \calO_t(\abs{v}^3) + \calO(\abs{t}^3)
       \end{align*}
       We want to replace $v$ with $\pi(\theta)$. Performing an expansion of this function around $e_2$ yields
       \begin{align*}
           \pi(\theta) = 0 + \widehat{\theta}+ (1-\theta_2)  \widehat{\theta} + \calO(d_\sph(e_2,\theta)^4) = \widehat{\theta}+ \calO(d_\sph(e_2,\theta)^3).
       \end{align*}
       Hence,
       \begin{align*}
           \pi^{-1}(t + \pi(\theta)) =& \exp(\alpha)e_2 - \left(\langle t, \widehat{\theta} \rangle + \tfrac{\abs{\widehat{\theta}}^2}{2}\right) \begin{bmatrix}
           t \\ 1
           \end{bmatrix}    
           +\left(1- \tfrac{\abs{t}^2}{2}-  \langle{t,\widehat{\theta}} \rangle \right)\begin{bmatrix} \widehat{\theta}\\ 0 \end{bmatrix} \\
           & +\calO(d_\sph(e_2,\theta)^3) + \calO(\abs{t}^3). \\
           =& \exp(\alpha)e_2 + \begin{bmatrix}\widehat{\theta} \\ 0 \end{bmatrix}-\langle t, \widehat{\theta}\rangle \begin{bmatrix} t+ \widehat{\theta} \\ 1 \end{bmatrix} - \tfrac{\vert{\widehat{\theta}}\vert^2}{2} \begin{bmatrix} t \\ 1 \end{bmatrix} - \tfrac{\abs{t}^2}{2} \begin{bmatrix} \widehat{\theta} \\ 0 \end{bmatrix}
           \\
           &+ \calO(d_\sph(e_2,\theta)^3) + \calO(\abs{t}^3).
       \end{align*}
       This is to be compared to $\exp(\alpha)\theta$, i.e.
       \begin{align*}
           \exp(\alpha) \theta &= \exp(\alpha) e_2 + (\id + \alpha + \tfrac{1}{2}\alpha^2 + \calO(\abs{\alpha}^3))(\theta-e_2)
       \end{align*}
      Using \eqref{eq:tvsalpha}, we obtain
        \begin{align*}
            &(\id + \alpha + \tfrac{1}{2}\alpha^2)(\theta-e_2) \\
            &= \begin{bmatrix} \widehat{\theta} \\ \theta_2 -1 \end{bmatrix} + \begin{bmatrix} t (\theta_2-1) \\ -\langle{t,\widehat{\theta}}\rangle \end{bmatrix}  - \tfrac{\abs{t}^2}{2} \begin{bmatrix} \widehat{\theta} \\ \theta_2 -1 \end{bmatrix} + \tfrac{1}{2} \langle{\widehat{\theta},\alpha}\rangle \begin{bmatrix}\alpha \\ 0 \end{bmatrix} + \calO(\abs{\alpha}^3) \\
            &= \begin{bmatrix}\widehat{\theta} \\ 0 \end{bmatrix} + (\theta_2-1) \begin{bmatrix} t \\ 1 \end{bmatrix} -\langle t, \widehat{\theta} \rangle e_2 - \tfrac{\abs{t}^2}{2} \begin{bmatrix} \widehat{\theta} \\ \theta_2 -1 \end{bmatrix} + \tfrac{1}{2} \langle{\widehat{\theta},\alpha}\rangle \begin{bmatrix}\alpha \\ 0 \end{bmatrix} + \calO(\abs{\alpha}^3)
        \end{align*}
        Remembering that $\theta_2-1 = \tfrac{\abs{\widehat{\theta}}^2}{2} + \calO(d_\sph(\theta,e_2)^4) $, we hence get
        \begin{align*}
            \chi_{\exp(\alpha)}(\theta) - \exp(\alpha)\theta =& -\langle{t,\widehat{\theta}\rangle} \begin{bmatrix} t + \widehat{\theta} \\ 0 \end{bmatrix}  + \tfrac{1}{2} \langle \widehat{\theta},\alpha \rangle \begin{bmatrix} \alpha \\ 0 \end{bmatrix} - \tfrac{\abs{t}^2}{2} (\theta_2-1)e_2 \\
            &+ \calO( d_\sph(e_2,\theta)^3) +\mathcal{O}(\abs{t}^3)
        \end{align*}
        We may safely insert a term  $-\langle t, \widehat{\theta}\rangle (\theta_2-1)e_2$, since it is of order $3$ in $d_\sph(e_2, \theta)$. Also, since $t \in \mathrm{span}(e_0,e_1)$, $\langle t, \widehat{\theta} \rangle =  \langle t, \theta \rangle$. The proof is finished.
    \end{proof}
    
    We see that just as we claimed in the main text, the error term gets particularly small for $\theta$ perpendicular to $t$ (at least for small $t$), meaning that it is small perpendicular to the translation direction. This is in contrast to the $\PY$-induced distortion, which is smallest along the direction of the rotation. The expression as an additional sweet-spot around $\widehat{\theta} \approx-t$, since the first two error terms vanish there. In fact, this can be precisized.
    
    \begin{prop}
        For $\omega = \pi^{-1}(-t)$, $\chi_\alpha(\omega)=\psi_\alpha(\omega)$.
    \end{prop}
    \begin{proof}
        It is a straighforward calculation to show that
        \begin{align*}
            \chi_\alpha(\omega) = \pi^{-1}(\pi(\omega)+t)= \pi^{-1}(-t+t)=e_2.
        \end{align*}
        Thus, we need to show that $\psi_\alpha (\omega)=\exp(\alpha)\omega= e_2$, or equivalently $-t = \pi(\exp(-\alpha)e_2)$. To prove this, it suffices to show that 
        \[\langle e_j,\exp(-\alpha)e_2\rangle=-\langle e_j,\exp(\alpha)e_2\rangle\] for $j=0,1$ and $\langle e_2,\exp(-\alpha)e_2\rangle = \langle e_2,\exp(\alpha)e_2\rangle$. For this, notice that $\alpha^n e_2$ is in $\mathrm{span}(e_2)$ for $n$ even and in $\mathrm{span}(e_0,e_1)$ for $n$ odd (this can for instance be shown via induction). This has the consequence that for $j=0,1$
        \begin{align*}
            \langle{e_j,\exp(-\alpha) e_2} \rangle &= \sum_{n \in \N} \tfrac{1}{n!} \langle{e_j,(-\alpha)^n e_2}\rangle \\
            &=\sum_{n  \text{ odd} }  \tfrac{1}{n!}\cdot (-1)^n \langle{e_j,\alpha^n e_2}\rangle \\
            &= \sum_{n  \text{ odd} }  -\tfrac{1}{n!}\cdot \langle{e_j,\alpha^n e_2}\rangle  \\
            &= -\sum_{n  \in \N }  \tfrac{1}{n!} \langle{e_j,\alpha^n e_2}\rangle \\
            &= - \langle{e_j,\exp(\alpha)e_2}\rangle, 
        \end{align*}
        and similarly $\langle e_2,\exp(-\alpha)e_2\rangle = \langle e_2,\exp(\alpha)e_2\rangle$. The proof is finished.
    \end{proof}
    
    Let us finally argue that the error becomes particularly small on a great circle perpendicular to the rotation direction.
    
\begin{prop}
    On the set $G=\{\theta \, \vert \,2 \langle{\pi(\theta),t}\rangle + \abs{t}^2=0\}\cap \{\theta \, \vert \, \theta_2>0 \}$,
    \begin{align*}
        \chi_\alpha(\theta) - \psi_\alpha(\theta)  = \tfrac{1}{4}\theta_2\vert \alpha \vert^2  + \mathcal{O}(\abs{\alpha}^3),
    \end{align*}
\end{prop}

\begin{proof}
    For $\theta \in G$, $\abs{\pi(\theta)} = \abs{\pi(\theta)+t}$. This has the consequence
    \begin{align*}
        \pi^{-1}(\pi(\theta) +t) &= \tfrac{1}{\sqrt{1+ \abs{\pi(\theta)+t}^2}}\begin{bmatrix} \pi(\theta) +t\\ 1 \end{bmatrix} \\
        &= \tfrac{1}{\sqrt{1+ \abs{\pi(\theta)}^2}}\begin{bmatrix}  \pi(\theta)\\ 1 \end{bmatrix} + \tfrac{1}{\sqrt{1+ \abs{\pi(\theta)}^2}}\begin{bmatrix}  t\\ 0 \end{bmatrix} 
        \\&= \pi^{-1}(\pi(\theta)) + \tfrac{1}{\sqrt{1+ \abs{\pi(\theta)}^2}}\begin{bmatrix}  t\\ 0 \end{bmatrix}
    \end{align*}
    We have $\sqrt{1+\abs{\pi(\theta)}^2}=\theta_2^{-1}$. This, together with \eqref{eq:tvsalpha}, implies
    \begin{align*}
        \pi^{-1}(\pi(\theta) + t) = \theta + \theta_2 \cdot  \alpha e_2 + \mathcal{O}(\abs{\alpha}^3).
    \end{align*}
    Since $\exp(\alpha) = 1+ \alpha + \tfrac{1}{2} \alpha^2 + \mathcal{O}(\abs{\alpha}^3)$, we obtain, using the notation of the previous proof,
    \begin{align*}
        &\chi_\alpha(\theta) - \psi_\alpha(\theta) \\
        &= -\alpha \begin{bmatrix} \widehat{\theta} \\ 0 \end{bmatrix} - \tfrac{1}{2}\alpha^2 \theta + \mathcal{O}(\abs{\alpha}^3) \\
        &=\langle t, \widehat{\theta}\rangle e_2 - \tfrac{1}{2} \abs{\alpha}^2 \theta_2 e_2 - \tfrac{1}{2}\begin{bmatrix}\alpha \\ 0 \end{bmatrix} \langle{t, \widehat{\theta}}\rangle + \mathcal{O}(\abs{\alpha}^3).
    \end{align*}
    Now we use that (up to higher order terms) $\abs{\alpha}^2 = \abs{t}^2$ and $\langle{t,\widehat{\theta}}\rangle = \theta_2 \langle t,\pi(\theta)\rangle= - \tfrac{1}{2} \theta_2\abs{t}^2 $ to see that the first two terms of the above vanish. Hence, the error is equal to than
    \begin{align*}
        \tfrac{1}{2}\vert{\langle{t, \widehat{\theta}}\rangle}\vert   + \mathcal{O}(\abs{\alpha}^3) =  \tfrac{1}{4}\theta_2\vert t\vert^2  + \mathcal{O}(\abs{\alpha}^3),
    \end{align*}
    which was the claim.
\end{proof}

Note that due to geometric properties of $\pi$, the equation $2 \langle{\pi(\theta),t}\rangle + \abs{t}^2=0$ indeed defines a great circle. The proposition shows that \emph{ near the equator} on this great circle $\theta_2\approx 0$, the error is very small. Note that this claim is only of academic interest for us, since points near the equator correspond to points very far away from the principal point, i.e. points that most often will not be captured on an image. We still chose to include it for completeness.

\newpage
\section{Proofs} \label{sec:proofs}
Here, we present the detailed proofs that were left out of the main text.

\subsection{Proposition  \ref{prop:singularSets}}\label{sec:proof_singularSets}

\begin{proof}[Proof of Proposition \ref{prop:singularSets}]
    Let $H \in \R^{3,3}$ be the representation of $\tau$ in homogenous coordinates, as in Remark \ref{rem:homography}. We are concerned with the solutions of the equation $\tau(\pi(x))=\pi(\rho(x))$. The latter is exactly the case when for some $\lambda \in \R$,
\begin{equation}\label{eq:translrigid}
\lambda Hx = \rho(x) =  Rx + v \Leftrightarrow (\lambda H - R)x = v.
\end{equation}
where $R$ and $v$ are defined through $\rho(x)=Rx+v$. Equation \eqref{eq:translrigid} has a unique solution as long as $\lambda H-R$ is invertible. The $\lambda$ for which this is not the case are the real eigenvalues of $H^{-1}R$. Since $H^{-1}R$ is a real $(3\times 3)$ matrix, it always has at least one eigenvalue, say $\lambda_0$. Since $H^{-1}R$ is not the identity, an eigenvalue $\lambda$ cannot have (geometric) multiplicity three. As for the case of multiplicity $2$, we have the following:

\underline{Claim:} If there exists a $\lambda$ such that $M_\lambda= \lambda H - R$ has rank $1$, $\mathrm{ker}(M_\lambda) = \mathrm{span}(e_0,e_1).$

\emph{Proof.} If $R$ is the identity, the claim follows immediately from the spectral theory for $H$. So assume that $R$ is not the identity. $R$ must then have a $2$-dimensional invariant subspace $W$.  Let $V = \mathrm{ker} M_\lambda$ be the at least two-dimensional kernel $V$. Since both the spaces are two-dimensional, $V \cap W$ is non-trivial and  both $R$ and $M_\lambda$-invariant. This leaves us with two cases.

\underline{Case 1: $R$ acts as minus the identity on $W$.} Then, we have proven the existence of a $v\in \R^3$ with $M_\lambda v =0$ and $Rv=-v$. This has the consequence that $\lambda Hv = Rv =-v$, so that $v$ is an eigenvector of $H$, and thus, in $\mathrm{span}(e_0,e_1)$. Furthermore, $\lambda = - 1$. This implies that for the other vector $w \in \mathrm{ker} M_\lambda$, $\abs{w}=\abs{Rw}= \abs{\lambda Hv} = \abs{Hw}$, which, by a direct computation, proves that also that vector $w$ is in $\mathrm{span}(e_0,e_1)$. Hence, $V=W=\mathrm{span}(e_0,e_1)$, which was to be proven.

\underline{Case 2: $W$ is irreducible.} In this case, we must conclude that $V=W$ -- otherwise, $W \cap V$ would be an $R$-invariant one-dimensional subspace of $W$. In other words, $\lambda H v = Rv$ for all $v \in W$. This in particular implies that $\lambda \neq 0$. Now let  $v_0, v_1,v_2$ of $\R^3$ be an orthonormal basis of $\R^3$ so that $Rv_0=v_0$, and consequently $\mathrm{span}(v_1,v_2)=V$. We have 
\begin{align*}
    \lambda \langle{v_i, H^Tv_0} \rangle= \lambda\langle{Hv_i,v_0}\rangle = \langle Rv_i, v_0 \rangle = \langle v_i, R^Tv_0 \rangle =0 , \quad i=1,2.
\end{align*}
Hence, $H^Tv_0$ is orthogonal to both $v_1$ and $v_2$, and hence, a multiple of $v_0$. $v_0$ is thus an eigenvector of $H^T$. Since $t$ is non-zero, at least one if its components are, WLOG $t_0\neq 0$. Then, $v_0$ being an eigenvector of $H^T$ implies that $v_0 \in \mathrm{span}(e_1,e_2)$, say $v_0=\cos(\vartheta)e_2 + \sin(\vartheta)e_1$.  Now, since $W$ is the orthogonal complement of $v_0$, $e_0$ must then be in $V=W$. This has the consequence
\begin{align}
    1=\abs{e_0}^2=\abs{Re_0}^2 = \abs{\lambda H e_0}^2 = \abs{\lambda}^2.
\label{eq:widerspruch1}
\end{align}
However, since also $u=-\sin(\vartheta)e_2 +\cos(\vartheta)e_1$ is orthogonal to $v_0$ and thus a member of $W=V$, we obtain
\begin{align}
    1= \abs{u}^2 = \abs{Ru}^2=\abs{\lambda Hu}^2 = \abs{\lambda} ^2(1+\sin(\vartheta)^2\abs{t}^2) \label{eq:widerspruch2}.
\end{align}
Equations \eqref{eq:widerspruch1} and \eqref{eq:widerspruch2} can only be true simultaneously if $\sin(\vartheta)=0$, i.e. that $\theta$ is a multiple of $\pi$, and $v_0=e_2$. $V=W$ is as the orthogonal complement of $v_0$ then equal to $\mathrm{span}(e_0,e_1)$, and the claim has been proven. ($\square$)

The above leaves us with three cases concerning the solveability of \eqref{eq:translrigid}:
\begin{enumerate}
    \item $\lambda H-R$ is regular. Then, the equation has the unique solution $(\lambda H-R)^{-1}v$.
    \item $\lambda H- R$ has rank $2$. If $v$ lies in the range of $\lambda H- R$. Then, the solutions of \eqref{eq:translrigid} form a one-dimensional subspace of $\R^3$. If not, the equation has no solutions.
    \item $\lambda H - R $ has rank $1$. Then, the kernel of $\lambda H-R$ is equal to $\mathrm(e_0,e_1)$. If $v$ is in the range of $(\lambda H-R)$, the set of solution is then a plane parallel to $\mathrm{span}(e_0,e_1)$. Otherwise, the solution-set is empty. In this case, $R$ must further be the identity or $\mathrm{diag}(-1,-1,1)$
\end{enumerate}
The $\lambda$ for which cases 2 and 3 occur are exactly the eigenvalues $M$ of $H^{-1}R$. We conclude that $S$ is given by the union of the curve
\begin{align*}
    \gamma: \R \backslash M  \to \R^3, \quad \lambda \mapsto (\lambda H-R)^{-1} v
\end{align*} 
and possibly sets of the form $w_0 + \ker(\lambda H-R)$ for $\lambda \in M$. The structures of these sets are described by cases 2 and 3 above, and exactly corresponds to the structure claimed in the proposition. The proof is finished.
\end{proof}

\begin{figure*}
    \centering
    \includegraphics[width=.3\textwidth]{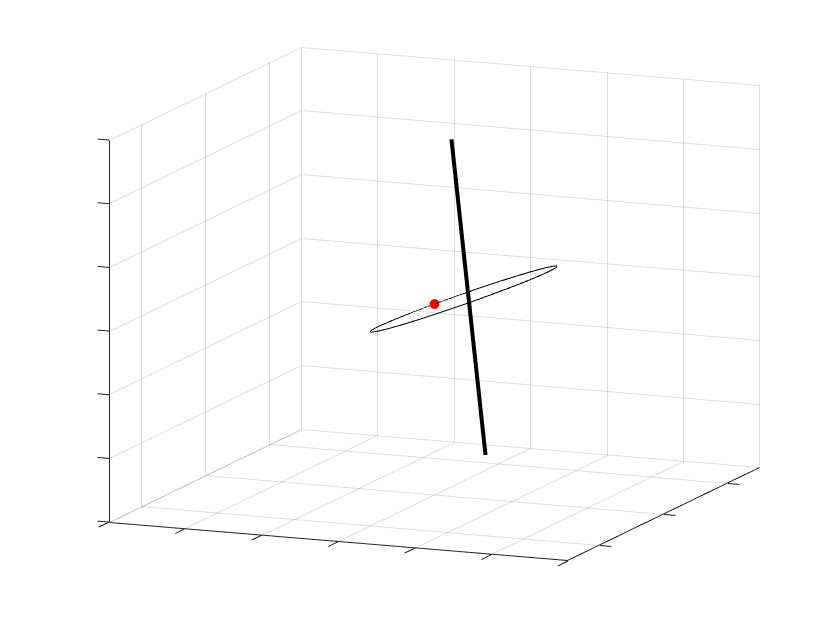}\includegraphics[width=.3\textwidth]{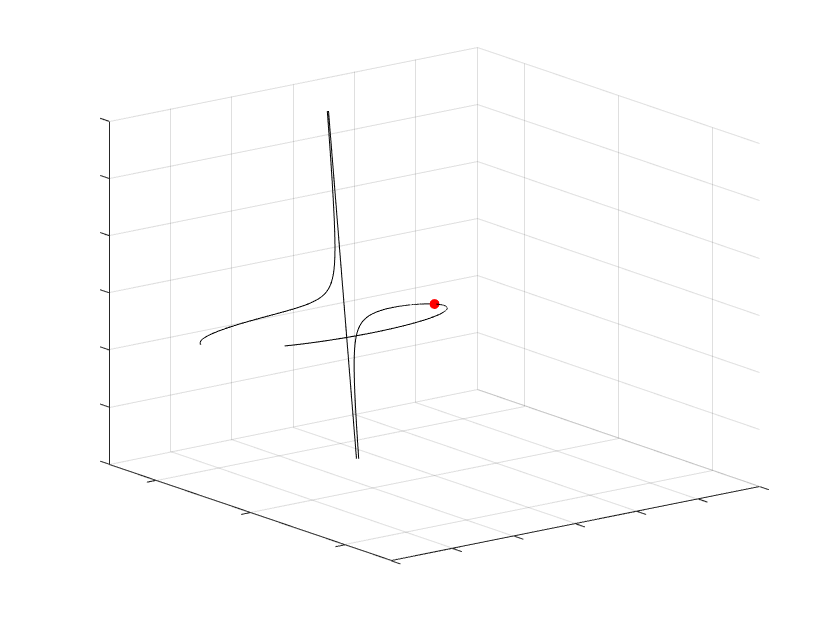}\includegraphics[width=.3\textwidth]{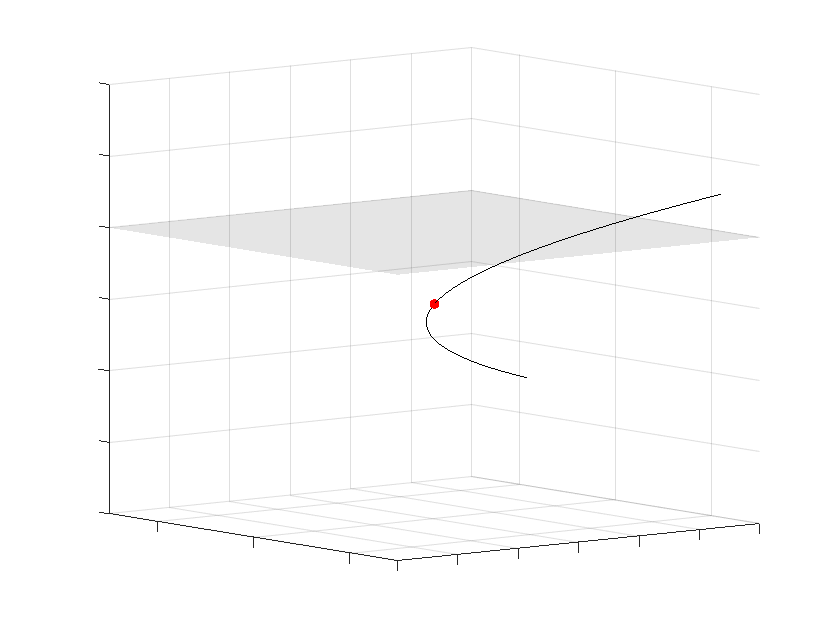}
    \caption{Three possible appearances of the $S$-set. The dot indicates the camera position.}
    \label{fig:SingSets}
\end{figure*}

An illustration of three possible $S$-sets can be seen in Figure~\ref{fig:SingSets}

\subsection{Proposition \ref{prop:HvsSO}}\label{prop:proof_HvsSO}

\begin{proof}[Proof of Proposition \ref{prop:HvsSO}]
    1. Let $\h_i$, $i=0,1$, be homographies with $\h_i \circ \pi = \pi \circ R_i$. Then we have for $x\in \R^3$ arbitrary
    \begin{equation*}
        (\h_0 \circ \h_1) (\pi(x)) = \h_0 \circ(\h_1 \circ \pi)(x)=(\h_0 \circ \pi) (R_1x) = \pi ( R_0R_1).
    \end{equation*}
    This proves $\h_0 \circ \h_1 \in \Hm$, and also that composition of the homographies is the same as multiplying the representing matrices. This proves that $\Hm$ is a subgroup of the diffeorphisms, and the map from a homography $h$ to its representing matrix $R$ is an isomorphism between groups.
    
    2. Let $\h \in \Hm$ with associate rotation $R \in \SO(3)$ be arbitrary. We need to prove that $\Phi(g \circ \h) = (\phi g) \circ \h$. Let us first note that $ (g \circ \h)\circ \pi=  (g \circ \pi) \circ R$. Applying the $\SO(3)$-equivariant network $\Psi$ to that relation yields
    \begin{align*}
        \Psi( (g \circ \h) \circ \pi) = \Psi ( (g \circ \pi) \circ R) = \Psi((g\circ \pi)) \circ R.
    \end{align*}
    We have $\Psi((g\circ \pi)) = \Phi(g) \circ \pi$. The above, together with $\h = \pi \circ R$, hence implies
    \begin{align*}
        \Phi (g \circ \h) \circ \pi = \Phi(g) \circ \pi \circ \h  = \Phi(g) \circ \h \circ \pi,
    \end{align*}
    which yields the claim.

\end{proof}

\subsection{Proposition  \ref{prop:pitchyawnegative}}\label{sec:proof_pitchyawnegative}

\begin{proof}[Proof of Prop. \ref{prop:pitchyawnegative}]
    If $G$ is a subgroup containing pitches and yaws, it must contain all matrices of the form
    \begin{align*}
        R= R_{x_0}(\alpha) R_{x_1}(\beta) R_{x_0}(\gamma),
    \end{align*}
    where $R_{x_0}(\alpha)$ and $R_{x_0}(\gamma)$ are rotations about the ${x_0}$-axis (i.e. pitches) and $R_{x_1}(\beta)$ a rotation about the ${x_1}$-axis (i.e. a yaw).  A classical result of differential geometry (the existence of \emph{Euler angles}) however states that each rotation $R$ can be written in that form. Hence, such a subgroup cannot be proper.  
\end{proof}

    \subsection{Transforming images on \texorpdfstring{$\mathbb{P}^2$}{P2} to images on \texorpdfstring{$\PY$}{PY}.} \label{sec:proofs_warp}
    
    Let us derive a recalculation formula between points $x\in \mathbb{P}^2$,  $\alpha\in \PY$ and $\theta \in \sph^2$. The result is easiest to formulate in  polar coordinates in $\PY$ and $\mathbb{P}^2$.
    
    \begin{prop} \label{prop:warp}
    Let $\phi_{\sph^2}$ and $\phi_{\mathrm{pol}}$   be the standard spherical and polar coordinate maps, i.e. 
    \begin{align*}
        \phi_{\sph^2}(\vartheta,\varphi) &= \begin{bmatrix} \sin(\vartheta)\cos(\varphi), \sin(\vartheta)\sin(\varphi), \cos(\vartheta)\end{bmatrix}^T, \quad \phi_{\mathrm{pol}}(r, \varphi) \\
        &= r\begin{bmatrix} \cos(\varphi), \sin(\varphi) \end{bmatrix}^T. 
    \end{align*}
    \begin{enumerate}
        \item \emph{Projection from $\PY$ to $\sph^2$}. For $r<\pi$, we have
        \begin{align*}
            \Phi(\phi_{\mathrm{pol}}(r,\varphi)) = \left(0, \phi_{\sph^2}(r,\varphi - \tfrac{\pi}{2})\right).
        \end{align*}
      \item \emph{Projection from $\PY$ to $\mathbb{P}^2$}. For $r<\pi/2$, we have
      \begin{align*}
          \pi(\Phi(\phi_{\mathrm{pol}}(r,\varphi)))= \phi_{\mathrm{pol}}(\tan(r),\varphi - \tfrac{\pi}{2}).
      \end{align*}
    \end{enumerate}
\end{prop}

  \begin{proof} 1. Let us denote $\alpha= \phi_{\mathrm{pol}}(r,\varphi)$ The curve $\gamma$ used to define $\Phi$ is by Lemma \ref{lem:geodesics} a geodesic with speed $\abs{\alpha}$. As such, it does not cross $\{\pm e_2\}$ as long as $\abs{\alpha}<\pi$. Since we assumed this, we thus have $\Phi(\alpha) = (0, \gamma(1))$. In order to compute the spherical coordinates of $\gamma(1)$, we again invoke Lemma \ref{lem:geodesics} to see that $\gamma'(0)=\alpha e_2$. Therefore, $\gamma$ is the great circle defined by intersecting $\sph^2$ with $\mathrm{span}({\alpha e_2,e_2})$. All points in this plane have the same polar angles as $\alpha e_2 = \begin{bmatrix} \alpha_1, -\alpha_0, 0 \end{bmatrix}$, i.e the polar angle of $\alpha$ minus $\tfrac{\pi}{2}$. Since $\gamma(1)$ is by at a geodesic distance from $e_2$ of $\abs{\alpha}$, and that this is equal to the azimuth angle, the proof is finished.
  
  2. This is now a trivial consequence of the definition of $\pi$
  \begin{align*}
      \pi(\Phi(\phi_{\mathrm{pol}}(r,\varphi)) &= \pi(\phi_{\sph^2}(r, \varphi - \tfrac{\pi}{2})) \\
      &= \tfrac{1}{\cos(r)} \begin{bmatrix} \sin(r) \cos(\varphi . \tfrac{\pi}{2}) 
      \\ \sin(r) \sin(\varphi - \tfrac{\pi}{2}) \end{bmatrix} \\
      &= \phi_{\mathrm{pol}}(\tan(r), \varphi-\tfrac{\pi}{2})  .
  \end{align*}
  \end{proof}

\section{Per-object results}
\label{sec:further_results} 
\label{sec:res_ep} 
\label{sec:results_per_object} 

Table~\ref{tab:occl_adds_perobj} shows the per-object results on \LMO{}, and Table~\ref{tab:lm_adds_perobj} shows the per-object results on \LM{}, in both cases training on the full training set.

\begin{figure}
    \centering
    \includegraphics[width=0.7\textwidth]{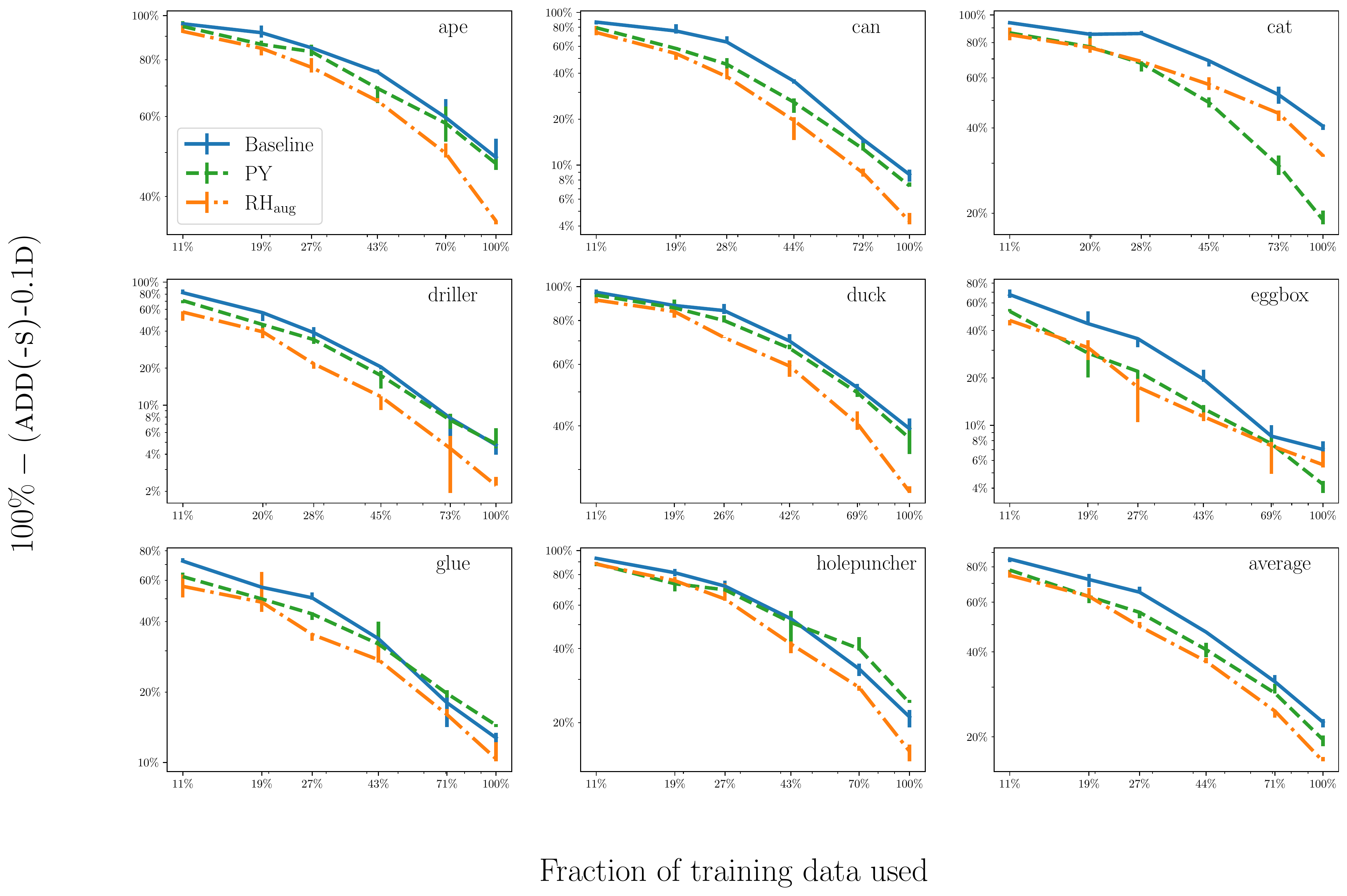}
    \caption{Detailed per-object results for the limited data experiments on \LMO{}.
    In all log-log plots, the line follows the median of three trials, whereas the error bars indicate the best and worst of the trials.
    The \add{} metric is inverted, such that the $y$-axis shows an error function $100 \% - \text{\add{}}$.
    The average plot from Figure~\ref{fig:limit_results} in the main article is also included for reference.
    }
    \label{fig:lmo_perobj}
\end{figure}

\begin{figure}
    \centering
    \includegraphics[width=0.95\textwidth]{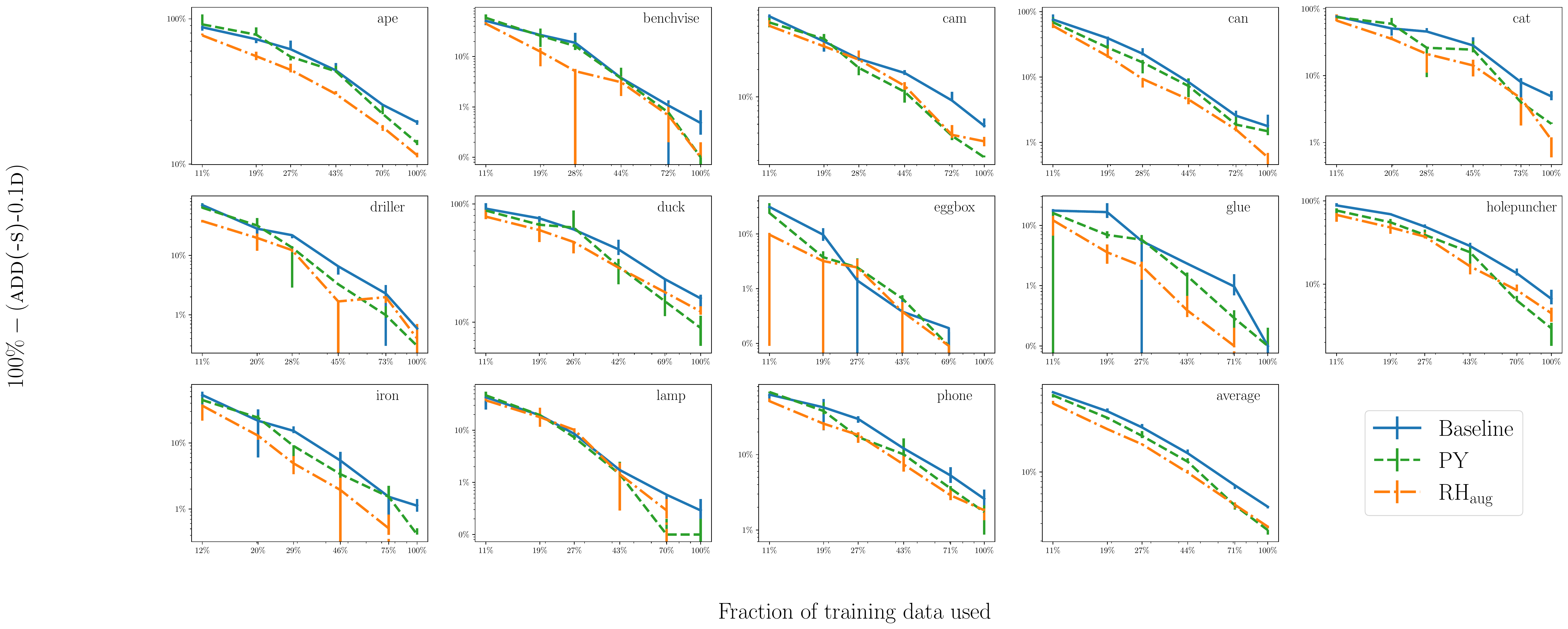}
    \caption{Detailed per-object results for the limited data experiments on \LM{}.
    In all log-log plots, the line follows the median of three trials, whereas the error bars indicate the best and worst of the trials.
    The \add{} metric is inverted, such that the $y$-axis shows an error function $100 \% - \text{\add{}}$.
    The average plot from Figure~\ref{fig:limit_results} in the main article is also included for reference.
    Note that some runs scored $100\%$ which is why some error bars drop away outside the plots.
    }
    \label{fig:lm_perobj}
\end{figure}

Figures \ref{fig:lmo_perobj} and \ref{fig:lm_perobj} show the per-object results for the limited data study on \LMO{} and \LM{} respectively.

\newcommand{\rottableheader}[1]{\rotatebox[origin=c]{70}{#1}}

\begin{table*}
    \centering
    \small
    \scalebox{.93}{
\begin{tabular}{|l||c|c|c|c||c|c|c|c|}
    \hline
    \multirow{2}{*}[-.3em]{Object}
        & \multicolumn{4}{c||}{$\phi=0$}
        & \multicolumn{4}{c|}{$\phi=3$}

        \bigstrut
        \\
     \cline{2-9}
         &
         Original
         &
         Baseline
         &
         $\PY$
         &
         $\Hm_{\mathrm{aug}}$
         &
         Original
         &
         Baseline
         &
         $\PY$
         &
         $\Hm_{\mathrm{aug}}$

        \bigstrut
                 \\
    \hline
    ape
       & \result{56.57} 
       & \result{51.21} 
       & \newcorrectcell \result{52.73} 
       & \result{64.65} 
       & \result{59.39} 
       & \result{55.25} 
       & \result{54.85} 
       & \resultbf{65.15} 
       \bigstrut
       \\
    \hline
    can
       & \result{91.12} 
       & \result{91.32} 
       & \newcorrectcell \result{92.68} 
       & \result{95.71} 
       & \result{93.27} 
       & \result{94.44} 
       & \result{93.66} 
       & \resultbf{96.68} 
       \bigstrut
       \\
    \hline
    cat
       & \result{68.58} 
       & \result{59.46} 
       & \newcorrectcell \result{81.07} 
       & \result{68.09} 
       & \result{79.78} 
       & \result{78.99} 
       & \resultbf{84.34} 
       & \result{70.17} 
       \bigstrut
       \\
    \hline
    driller
       & \result{95.64} 
       & \result{95.25} 
       & \newcorrectcell \result{95.15} 
       & \result{97.77} 
       & \result{97.77} 
       & \result{97.67} 
       & \result{96.90} 
       & \resultbf{98.55} 
       \bigstrut
       \\
    \hline
    duck
       & \result{65.31} 
       & \result{60.73} 
       & \newcorrectcell \result{63.12} 
       & \result{74.06} 
       & \result{72.71} 
       & \result{65.10} 
       & \result{69.27} 
       & \resultbf{76.56} 
       \bigstrut
       \\
    \hline
    eggbox
       & \result{93.46} 
       & \result{92.96} 
       & \newcorrectcell \result{95.77} 
       & \result{94.37} 
       & \resultbf{96.18} 
       & \result{95.47} 
       & \resultbf{96.18} 
       & \result{95.47} 
       \bigstrut
       \\
    \hline
    glue
       & \result{85.15} 
       & \result{87.25} 
       & \newcorrectcell \result{85.55} 
       & \result{89.62} 
       & \result{90.80} 
       & \result{90.01} 
       & \result{89.49} 
       & \resultbf{91.59} 
       \bigstrut
       \\
    \hline
    holepuncher
       & \result{76.53} 
       & \result{78.90} 
       & \newcorrectcell \result{75.94} 
       & \result{84.71} 
       & \result{81.95} 
       & \result{81.07} 
       & \result{78.01} 
       & \resultbf{85.21} 
       \bigstrut
       \\
    \hhline{|=========|}
    Average
       & \result{79.04} 
       & \result{77.44} 
       & \newcorrectcell \result{80.43} 
       & \result{83.58} 
       & \result{83.98} 
       & \result{81.94} 
       & \result{82.91} 
       & \resultbf{84.92} 
       \bigstrut
       \\
    \hline
\end{tabular}

    }
    \caption{Detailed per-object 6D object pose results according to the \add{} metric, with \EP{} on \LMO{}, and train/test split as in \cite{bukschat2020efficientpose}.
    Results are the median over three reruns.
    The results in the final row correspond to Table~\ref{tab:occl_aggregated} in the main article and are the median (across the reruns) of the average \add{} across all objects.
    The ``Baseline'' models are retrained and evaluated by us, after some minor modifications, while ``Original'' refers to the results reported by \cite{bukschat2020efficientpose}.
    Model size is controlled by the $\phi$ parameter.}
    \label{tab:occl_adds_perobj}
\end{table*}

\begin{table*}
    \centering
    \small
\begin{tabular}{|l||r|r|r|r|r|}
    \hline
     &  & \multicolumn{1}{l|}{Original} & &  & \\
    Object & \multicolumn{1}{l|}{Original} & \multicolumn{1}{l|}{($\phi=3$)} & \multicolumn{1}{l|}{Baseline} & \multicolumn{1}{l|}{$\PY$} & \multicolumn{1}{l|}{$\Hm_\mathrm{aug}$} \bigstrut\\
    \hline
    ape
        & \result{87.71}
        & \resultbf{89.43}
        & \result{80.57}
        & \newcorrectcell \result{86.00}
        & \result{88.48}
    \bigstrut\\
    \hline
    benchvise
        & \result{99.71}
        & \result{99.71}
        & \result{99.52}
        & \newcorrectcell \resultbf{99.90}
        & \resultbf{99.90}
    \bigstrut\\
    \hline
    cam
        & \result{97.94}
        & \resultbf{98.53}
        & \result{95.20}
        & \newcorrectcell \result{97.84}
        & \result{96.76}
    \bigstrut\\
    \hline
    can
        & \result{98.52}
        & \resultbf{99.70}
        & \result{98.23}
        & \newcorrectcell \result{98.52}
        & \result{99.41}
    \bigstrut\\
    \hline
    cat
        & \result{98.00}
        & \result{96.21}
        & \result{95.11}
        & \newcorrectcell \result{98.10}
        & \resultbf{98.90}
    \bigstrut\\
    \hline
    driller
        & \resultbf{99.90}
        & \result{99.50}
        & \result{99.41}
        & \newcorrectcell \result{99.70}
        & \result{99.60}
    \bigstrut\\
    \hline
    duck
        & \result{90.99}
        & \result{89.20}
        & \result{84.13}
        & \newcorrectcell \resultbf{91.08}
        & \result{87.70}
    \bigstrut\\
    \hline
    eggbox
        & \resultbf{100.00}
        & \resultbf{100.00}
        & \resultbf{100.00}
        & \newcorrectcell \resultbf{100.00}
        & \resultbf{100.00}
    \bigstrut\\
    \hline
    glue
        & \resultbf{100.00}
        & \resultbf{100.00}
        & \result{99.90}
        & \newcorrectcell \result{99.90}
        & \resultbf{100.00}
    \bigstrut\\
    \hline
    holepuncher
        & \result{95.15}
        & \result{95.72}
        & \result{93.34}
        & \newcorrectcell \resultbf{97.05}
        & \result{95.53}
    \bigstrut\\
    \hline
    iron
        & \result{99.69}
        & \result{99.08}
        & \result{98.88}
        & \newcorrectcell \result{99.59}
        & \resultbf{100.00}
    \bigstrut\\
    \hline
    lamp
        & \resultbf{100.00}
        & \resultbf{100.00}
        & \result{99.71}
        & \newcorrectcell \result{99.90}
        & \resultbf{100.00}
    \bigstrut\\
    \hline
    phone
        & \result{97.98}
        & \resultbf{98.46}
        & \result{97.41}
        & \newcorrectcell \result{98.27}
        & \result{98.17}
    \bigstrut\\
    \hhline{|======|}
    Average
        & \result{97.35}
        & \result{97.35}
        & \result{95.56}
        & \newcorrectcell \resultbf{97.43}
        & \result{97.21}
    \bigstrut\\
    \hline
    \end{tabular}%

    \caption{Detailed per-object 6D object pose results according to the \add{} metric, with \EP{} on \LM{}.
    Results are the median over three reruns.
    The results in the final row correspond to Table~\ref{tab:lm_aggregated} in the main article and are the median (across the reruns) of the average \add{} across all objects.
    The ``Baseline'' model is retrained and evaluated by us, after some minor modifications, while ``Original'' refers to the results reported by \cite{bukschat2020efficientpose}. All models are of type $\phi=0$, except the second result column which we include for reference.}
    \label{tab:lm_adds_perobj}
\end{table*}

\section{Visual localization experiments}\label{sec:appendix_localization}
Here we present experiments on visual localization that did not fit into the main paper.
The results are less striking than the 6D object pose estimation results, but still indicate that
improving rotation equivariance can be a useful tool for localization tasks.
Section~\ref{sec:exp_dsac} describes experiments using DSAC* Tiny \cite{brachmann-pami-2021} on 7-Scenes and Section~\ref{sec:exp_pixloc} describes experiments using Pixloc \cite{sarlin2021feature} on CMU Seasons.

\subsection{DSAC* Tiny on 7-Scenes} \label{sec:exp_dsac}
In this section we present experiments using DSAC* Tiny on the 7-Scenes dataset.

\subsubsection{Model. } Our experiments build on DSAC* Tiny 
 \cite{brachmann-pami-2021} 
 at \cite{dsac_github}.
 DSAC* is an end-to-end learning method. The first step of the DSAC* pipeline for RGB-input is using a CNN to predict scene coordinates for each pixel in the input image in order to obtain a dense map of 2D-3D correspondences, which  are then fed into DSAC \cite{brachmann2018dsac} (a differentiable modification of RANSAC \cite{fischler-bolles-cacm-1981}) with a PnP solver for estimation of the camera pose. At test time, standard RANSAC is used instead of DSAC. Lastly, the estimated camera pose is iteratively refined to maximize the number of inlier correspondences.
DSAC* Tiny is a small version of DSAC*, requiring only 4MB of memory. We work with a small model as we expect that equivariance will matter more when the model has low representational power. For more details we refer the reader to \cite{brachmann-pami-2021}. 
We make some minor modifications of the code as follows:

\begin{figure*}
    \centering
    \begin{tikzpicture}[node distance=.6cm and .6cm]
        \node[rotate=90] (in) {Input};
        \node[3x3, right=of in] (conv1) {$1 \to 32$};
        \node[3x3s2, right=of conv1] (conv2) {$32 \to 64$};
        \node[3x3s2, right=of conv2] (conv3) {$64 \to 128$};
        \node[3x3s2, right=of conv3] (conv4) {$128 \to 128$};
        \node[3x3, right=of conv4] (r1conv1) {$128 \to 128$};
        \node[1x1, right=of r1conv1] (r1conv2) {$128 \to 128$};
        \node[3x3, right=of r1conv2] (r1conv3) {$128 \to 128$};
        \node[3x3, right=of r1conv3] (r2conv1) {$128 \to 128$};
        \node[1x1, right=of r2conv1] (r2conv2) {$128 \to 128$};
        \node[3x3, right=of r2conv2] (r2conv3) {$128 \to 128$};
        \node[1x1, right=of r2conv3] (r3conv1) {$128 \to 128$};
        \node[1x1, right=of r3conv1] (r3conv2) {$128 \to 128$};
        \node[1x1, right=of r3conv2] (r3conv3) {$128 \to 128$};
        \node[1x1, right=of r3conv3] (fc1) {$128 \to 128$}; 
        \node[1x1, right=of fc1] (fc2) {$128 \to 128$};
        \node[1x1, right=of fc2] (fc3) {$128 \to 3$};
        \node[rotate=90, rectangle, on grid, right=of fc3] (out) {Output};
        \draw[bend left,->,thick] ($(conv4.east)+(0.35,0)$) to node [auto] {$+$} ($(r2conv1.east)-(0.39,0)$);
        \draw[bend left,->,thick] ($(r1conv3.east)+(0.35,0)$) to node [auto] {$+$} ($(r3conv1.east)-(0.39,0)$);
        \draw[bend left,->,thick] ($(r2conv3.east)+(0.35,0)$) to node [auto] {$+$} ($(fc1.east)-(0.39,0)$);
    \end{tikzpicture}
    \caption{The CNN in DSAC* Tiny \cite{brachmann-pami-2021}. The network takes a single grayscale image as input and outputs 3D coordinates corresponding to 2D pixels in a subsampled image. The \textcolor{green}{green} layers consist of $3\times3$-convolutions and the \textcolor{blue}{blue} layers of $1\times1$-convolutions. Between consecutive convolution layers, there are ReLU activations. The notation in the boxes means ``{$\text{number of input channels}\to\text{number of output channels}$}''. Skip connections are shown by the arrows marked with $+$. The \ugreendotted{dotted} layers use convolution with stride $2$, while all others use stride $1$. Every $3\times3$-convolution uses padding of one pixel around the image.}
    \label{fig:dsactinylayout}
\end{figure*}

\paragraph{Pixel grid. } The CNN in the DSAC* Tiny model has the layout described in Figure~\ref{fig:dsactinylayout}. Of particular note is the fact that three layers use stride 2 convolutions, leading to a subsampling of the input image dimensions by a factor of around 8 (depending on the input image). When calculating the pixel coordinates corresponding to the predicted 3D coordinates one has to keep this fact in mind. The original DSAC* code does this by subsampling the original pixels, starting at pixel $[4,4]$ and incrementing by $8$ in the $x_0$ and $x_1$ directions until reaching the image size. However, since the convolutions are padded, we believe it to be more correct to start at pixel $[0,0]$ and thus implement this change.

\paragraph{Augmentation mask. } In \cite{brachmann-pami-2021}, the authors state that they see deteriorated performance when using data augmentation with the low capacity DSAC* Tiny model. Out of concern for this, we add an augmentation mask to the code, so that only those pixels in an augmented image which correspond to pixel coordinates inside the original image contribute to the loss function. 

\subsubsection{Data augmentation}

Except for the experiment with general rotational homographies ( $\Hm_\mathrm{aug}$), the augmentation procedure follows the one in \cite{brachmann-pami-2021}, which in turn is inspired by the augmentation procedure in \cite{li2020hierarchical}.

All experiments use color jitter data augmentation during both training phases, by which we mean randomly rescaling the brightness and contrast of an image within $\pm 10\%$. We also consider two types of augmentations.

\paragraph{Geometric data augmentation. }
The 'geometric data augmentation' is as described in \cite{brachmann-pami-2021}: We randomly rotate the images in the range of $\pm30$ degrees in the image plane and rescaling the image by a random factor between $\sfrac{2}{3}$ and $\sfrac{3}{2}$. The ground truth camera pose is altered according to the rotation and the camera calibration according to the rescaling. 

 For simplicity, we use geometric data augmentation only during the first training phase, which consists of 10 times more epochs than the second training phase. 

\paragraph{$\Hm$ data augmentation. } The $\Hm$ augmentation is conducted just as for \EP{} (see Section \ref{sec:impl_details_homo_aug}) 
It is only applied during the first training phase. 

\subsubsection{Transformation to \texorpdfstring{$\PY$}{PY}}
For the experiments called $\PY$, we transform the input image to the $\PY$-domain and retransform the CNN output to the original $\mathbb P^2$-domain before feeding the correspondences into DSAC. This approach was the simplest as the DSAC implementation bundled with DSAC* uses a dense prediction of 2D-3D correspondences relating every pixel in an (subsampled) image to a scene coordinate.

The transformation to the $\PY$-domain proceeds just as in the experiments with EfficientPose, see Section~\ref{sec:impl_details_py}. Transforming back is done similarily, for every pixel in the subsampled image we calculate its $\PY$-coordinates and bilinearly interpolate the scene coordinate from the output of the CNN which is in the $\PY$-domain. 

\subsubsection{Dataset.} We perform visual localization experiments on the 7-Scenes dataset
\cite{shotton2013scene}
consisting of collections of 
images of 7 different scenes, taken by a Kinect camera.
We discard the depth information and
estimate
the camera pose from a single RGB image. We use the standard train-test-split for 7-Scenes that is provided with the dataset.

\subsubsection{Training setup} We test five different training setups: 
 Baseline (only color jitter augmentation), $\PY$ (resampling the image to the $\PY$-domain, still only with color jitter augmentation) 
  Baseline+GeomAug, $\PY$+GeomAug,  (as Baseline/$\PY$, but additionally augmenting with image size changes and small rotations in the image plane) and $\Hm_\mathrm{aug}$ (Baseline + Geomaug + $\Hm$-augmentation). This is summarized in Table~\ref{tab:dsacstar_experiments}.  We follow \cite{brachmann-pami-2021} for hyperparameter choices and do not perform any hyperparameter optimization ourselves. 

We train all models on the same kind of cluster compute nodes as described in Section \ref{sec:ep_hardware}, but with a Singularity container running Debian 10, and with Pytorch 1.6 installed.
The initial training phase takes around 7 hours and the end-to-end training phase takes around 8 hours.
  
  \begin{table*}[htbp] 
    \centering
    \begin{tabular}{c|c|c|c|c|}
        &  Image & Image & & \\  
          Experiment &  domain &  scalings & Rolls (${}^{\circ}$) & Tilts (${}^{\circ}$)  \\ \hline \\
        Baseline &  $\mathbb P^2$ & None & None & None \\
        Baseline+GeomAug &  $\mathbb P^2$ & $[\sfrac{2}{3},\sfrac{3}{2}]$ & $[-30,30]$ & None \\
        $\Hm_\mathrm{aug}$ &  $\mathbb P^2$ & $[\sfrac{2}{3},\sfrac{3}{2}]$ & $[-30,30] $ & $[0,20]$ \\
        $\PY$ &  $\PY$ & None & None & None \\
        $\PY$+GeomAug &  $\PY$ & $[\sfrac{2}{3},\sfrac{3}{2}]$ & $[-30,30]$ & None       
    \end{tabular}
    \caption{The experiment settings for visual localization on 7-Scenes with DSAC* Tiny. Rolls and tilts are defined in Section \ref{sec:impl_details_homo_aug}. When an interval is specified, we sample a value in the interval for each training image in each epoch.}
   \label{tab:dsacstar_experiments}
\end{table*}
  
  \begin{table*}[!htbp]
  \centering
  \small
    \begin{tabular}{|c|c|c|c|c|c|}
    \hline
    Experiment & \multicolumn{1}{l|}{$\PY$+GeomAug} & \multicolumn{1}{l|}{Baseline+GeomAug} & \multicolumn{1}{l|}{$\Hm_\mathrm{aug}$} & \multicolumn{1}{l|}{Baseline} & \multicolumn{1}{l|}{$\PY$} \\
    \hline
    Pose within 5cm/5° & {\bf 71.6} \% & 69.5\% & 68.0\% & 65.5\% & 64.9\% \\
    \hline
    \end{tabular}%
    \caption{Results on 7-Scenes. The scores are the proportion of test images with pose error within 5cm and 5°.}
  \label{tab:dsacstar_summary}%
\end{table*}%
  
   \subsubsection{Results. } The accuracy averaged over the test images in all scenes is presented in Table~\ref{tab:dsacstar_summary}.  More detailed results for the individual scenes can be found in
   Table~\ref{tab:dsac_results}.
 Each reported result is from a single run of training (per scene). 
 It is clear from Table~\ref{tab:dsacstar_summary} that data augmentation gives a performance boost. The best performance is obtained by  $\PY$+GeomAug, which we speculate is a good trade-off between homogenizing the data by resampling it and still having diverse realistic training examples provided by the augmentation. Correspondingly, the performance of $\Hm_{\mathrm{aug}}$ is slightly worse compared to Baseline+GeomAug. This might be due to the low capacity of the DSAC* Tiny model, as the diversity of the training set is increased substantially with the added augmentations.
 \begin{table*}[htbp]
  \centering
  \tiny

       \begin{tabular}{|ll||r|r|r|r|r|}
       \cline{1-7}
    \multicolumn{2}{|l||}{\diagbox{Scene, Metric}{Experiment}}   & \multicolumn{1}{l|}{$\PY$+GeomAug} & \multicolumn{1}{l|}{Baseline+GeomAug} & \multicolumn{1}{l|}{$\Hm_\mathrm{aug}$} & \multicolumn{1}{l|}{Baseline} & \multicolumn{1}{l|}{$\PY$} \bigstrut\\
    \hline\hline
    chess  &  5cm/5° & \textbf{95.00\%} & 94.80\% & 93.90\% & 94.20\% & 94.50\% \bigstrut\\
\cline{2-7}          & med. err. (cm) & 1.916 & \textbf{1.791} & 1.900   & 1.910  & 2.022 \bigstrut\\
\cline{2-7}          & med. err. (°) & 0.646 & \textbf{0.632} & 0.659 & 0.655 & 0.685 \bigstrut\\
    \hline
    fire  &  5cm/5° & \textbf{89.40\%} & 88.10\% & 88.40\% & 73.50\% & 73.00\% \bigstrut\\
\cline{2-7}          & med. err. (cm) & \textbf{2.114} & 2.231 & 2.183 & 2.713 & 2.694 \bigstrut\\
\cline{2-7}          & med. err. (°) & 0.945 & 0.952 & \textbf{0.936} & 1.035 & 1.031 \bigstrut\\
    \hline
    heads  &  5cm/5° & 94.20\% & 93.50\% & \textbf{94.40\%} & 84.40\% & 82.60\% \bigstrut\\
\cline{2-7}          & med. err. (cm) & \textbf{1.189} & 1.266 & 1.260  & 1.941 & 1.866 \bigstrut\\
\cline{2-7}          & med. err. (°) & \textbf{0.867} & 0.876 & 0.894 & 1.256 & 1.264 \bigstrut\\
    \hline
    office   &  5cm/5° & \textbf{77.50\%} & 71.50\% & 65.90\% & 74.10\% & 73.70\% \bigstrut\\
\cline{2-7}          & med. err. (cm) & \textbf{3.179} & 3.469 & 3.842 & 3.246 & 3.393 \bigstrut\\
\cline{2-7}          & med. err. (°) & 0.924 & 1.042 & 1.134 & 0.894 & \textbf{0.890} \bigstrut\\
    \hline
    pumpkin   &  5cm/5° & 58.50\% & 58.50\% & 56.50\% & 57.60\% & \textbf{61.10\%} \bigstrut\\
\cline{2-7}          & med. err. (cm) & 4.242 & 4.186 & 4.292 & 4.136 & \textbf{4.056} \bigstrut\\
\cline{2-7}          & med. err. (°) & 1.099 & 1.112 & 1.106 & \textbf{1.094} & 1.138 \bigstrut\\
    \hline
    redkitchen   &  5cm/5° & \textbf{58.80\%} & 56.50\% & 57.10\% & 55.50\% & 53.60\% \bigstrut\\
\cline{2-7}          & med. err. (cm) & \textbf{4.412} & 4.526 & 4.519 & 4.571 & 4.718 \bigstrut\\
\cline{2-7}          & med. err. (°) & 1.468 & 1.511 & 1.549 & \textbf{1.361} & 1.379 \bigstrut\\
    \hline
    stairs  &  5cm/5° & 33.50\% & \textbf{36.70\%} & 34.50\% & 4.20\% & 1.60\% \bigstrut\\
\cline{2-7}          & med. err. (cm) & 7.080  & \textbf{6.468} & 6.926 & 27.317 & 32.446 \bigstrut\\
\cline{2-7}          & med. err. (°) & \textbf{1.372} & 1.496 & 1.544 & 3.622 & 4.485 \bigstrut\\
    \hline\hline
    Total   &  5cm/5° & \textbf{71.60\%} & 69.50\% & 68.00\% & 65.50\% & 64.90\% \bigstrut\\
    \hline
    \end{tabular}%

    \smallbreak
    \caption{Detailed per-scene results of the experiments with DSAC* Tiny on 7-Scenes. 5cm/5° is the percentage of test examples where the predicted pose is within 5cm and 5° from the ground truth. The other reported metrics are the median error over the test set in cm and degrees. The \emph{Total} row refers to the percentage of accurate test examples over all scenes.}
  \label{tab:dsac_results}%
\end{table*}%

\subsection{Pixloc on CMU Seasons} \label{sec:exp_pixloc}

In this section we present experiments using Pixloc on the CMU Seasons dataset.

\subsubsection{Model.}
 The idea of Pixloc is to use a U-net to learn feature maps and calculate them for query images. The same features are computed for reference images with known pose, the reference images are found by image retrieval. One then optimizes using Levenberg-Marquardt for the query pose giving the best match between the features of a 3D model projected into both the query and reference images. For details about the Pixloc model, we refer to \cite{sarlin2021feature}. Notably, we are using a revised version of the code made available by \cite{sarlin2021feature} in the github repository \cite{pixlocgithub} after the publication of the article.
 
 In the undistorted, $\PY$-warp and $\Hm$-augment experiments we undistort, warp and/or augment both the query and reference images and modify the corresponding cameras accordingly. In both the warp and augment cases we undistort the image prior to further processing. We also change which 3D points should be considered for projections into the images according to the new cameras. We implement the derivative for the warp projection to enable the Levenberg-Marquardt method used in Pixloc.

\subsubsection{Dataset.} We use the CMU Extended Seasons dataset \cite{Badino2011, Toft2020}. This dataset is collected using car-mounted cameras, producing images with a significant amount of radial distortion. \cite{sarlin2021feature} take the radial distortion into account, e.g. when projecting the 3D models to the image plane, by considering a radial polynomial distortion model. Notably, this radial polynomial model is already quite close to our map between $\mathbb{P}^2\to \PY$. However, they do not use their model to undistort the images before training. This means that the baseline model should perform similarly to the model trained on $\PY$-warped versions of the undistorted images. To more clearly examine the effect of our methods, we therefore include a version of the model where perform the undistortion. See Figure~\ref{fig:warp_cmu} for an example of how similar the original and $\PY$-warped images are in this case.   

We use the training-validation split supplied by the Pixloc authors. Similarly as in the \LM{} experiment, we train the model on a randomly subsampled fraction of the original data. The fractions we consider are $(4^{-k})_{k =0,1,2,\dots,5}$. Here we subsample the query images but let the reference image set stay intact, in order to let the distance in pose from the query to the nearest reference be similar in the training and test sets. 

\subsubsection{Training setup}  We use the same hyperparameters as \cite{sarlin2021feature}, except for the fact that we explicitly stop the training after 45k iterations or 200 epochs (which one is earlier depends on the data proportion) -- \cite{sarlin2021feature} suggest manually braking the training after the validation loss has converged at around ``30k-40k'' iterations. We find that simply breaking at 45k iterations or 200 epochs works well. Of note is that the batch size of 3 means that 45k iterations corresponds to going through $3\cdot 45000 = 135000$ query-ref pairs. This is lower than the total available amount in the full data case and in the $4^{-1}=0.25$ and $4^{-2}\approx 0.063$ proportion cases it means that the available data is not passed through many times during training, giving a possible explanation of the similar performance of these proportions to the full data case.

The models are trained on the same kind of cluster compute nodes as described in Section~\ref{sec:ep_hardware}, but with a Singularity container running Debian 10 with Python 3.9 and Pytorch 1.10. The training takes around 48 hours.

\subsubsection{Results.} 
\begin{figure}
   \begin{minipage}{0.7\linewidth}
   \centering
    Recall at ($25\si{cm}$, $2\si{\degree}$)
    \begin{tabular}{|c||c|c|c|c|}
    \hline
    \multirow{2}{*}{{\small URBAN}}
        & \multicolumn{2}{c|}{Non-augmented}
        & \multicolumn{2}{c|}{$\Hm_{\mathrm{aug}}$-augmented} \\

     \cline{2-5}
        & Undist. & $\PY$
        & Undist. & $\PY$ \\
        
   \hline  Data fraction $4^{-5}$  & \result{79.5} & {\result{\bf 80.5}} & \result{77.3} & \result{79.8}\\ 
   \hline  \phantom{Data fraction} $4^{-4}$  & \result{85.5} & {\bf\result{ \bf 86.4}} & \result{84.3} & \result{85.6}\\ 
   \hline  \phantom{Data fraction} $4^{-3}$  & \result{89.0} & {\bf\result{\bf 89.6}} & \result{89.0} & \result{88.9}\\ 
   \hline  \phantom{Data fraction} $4^{-2}$  & {\bf\result{\bf 90.3}}& \result{90.0} & \result{89.2} & \result{89.2}\\ 
   \hline  \phantom{Data fraction} $4^{-1}$  & {\bf\result{\bf 90.6}} & \result{90.3} & \result{89.3} & \result{89.0}\\ 
   \hline  \phantom{Data fraction} $1.0$  & {\bf\result{\bf91.1}} & \result{90.5} & \result{88.9} & \result{88.9} \\
    \hline
\end{tabular}%
    \begin{tabular}{|c||c|c|c|c|c|}
    \hline
    \multirow{2}{*}{\small{SUBURBAN}}
        & \multicolumn{2}{c|}{Non-augmented}
        & \multicolumn{2}{c|}{$\Hm_{\mathrm{aug}}$-augmented} \\

     \cline{2-5}
        & Undist. & $\PY$
        & Undist. & $\PY$ \\
        
     \hline
     Data fraction $4^{-5}$  & \result{69.6} & \result{\bf 72.1} & \result{69.0} & \result{71.7}\\ 
     \hline  \phantom{Data fraction} $4^{-4}$  & \result{77.3} & \result{78.3} & \result{78.2} & \result{\bf 79.9}\\ 
     \hline  \phantom{Data fraction} $4^{-3}$  & \result{82.0} & \result{82.1} & \result{\bf 82.6} & \result{82.5}\\ 
     \hline  \phantom{Data fraction} $4^{-2}$  & \result{82.6} & \result{\bf 83.3} & \result{ 83.2} & \result{82.7}\\ 
     \hline  \phantom{Data fraction} $4^{-1}$  & \result{83.1} & \result{\bf 83.4} & \result{ 82.4} & \result{81.9}\\ 
     \hline  \phantom{Data fraction} $1.0$  & \result{\bf 83.5} & \result{82.9} & \result{82.6} & \result{82.7} \\
     \hline
\end{tabular}%
    \begin{tabular}{|c||c|c|c|c|c|}
    \hline
    \multirow{2}{*}{\small{PARK}}
        & \multicolumn{2}{c|}{Non-augmented}
        & \multicolumn{2}{c|}{$\Hm_{\mathrm{aug}}$-augmented} \\

     \cline{2-5}
        & Undist. & $\PY$
        & Undist. & $\PY$ \\
        
\hline  Data fraction $4^{-5}$  & \result{49.9} & \result{\bf 52.7} & \result{50.5} & \result{51.7}\\ 
\hline  \phantom{Data fraction} $4^{-4}$  & \result{58.9} & \result{59.6} & \result{60.8} & \result{\bf 61.2}\\ 
\hline  \phantom{Data fraction} $4^{-3}$  & \result{63.1} & \result{63.4} & \result{64.2} & \result{\bf 64.7}\\ 
\hline  \phantom{Data fraction} $4^{-2}$  & \result{63.9} & \result{65.0} & \result{\bf 66.2} & \result{65.5}\\ 
\hline  \phantom{Data fraction} $4^{-1}$  & \result{\bf65.8} & \result{65.5} & \result{64.8} & \result{63.9}\\ 
\hline  \phantom{Data fraction} $1.0$  & \result{\bf 66.4} & \result{65.0} & \result{64.8} & \result{65.3}
        \\
     \hline
\end{tabular}%
   \end{minipage}
   \begin{minipage}{0.28\linewidth}
    \includegraphics[width=\textwidth]{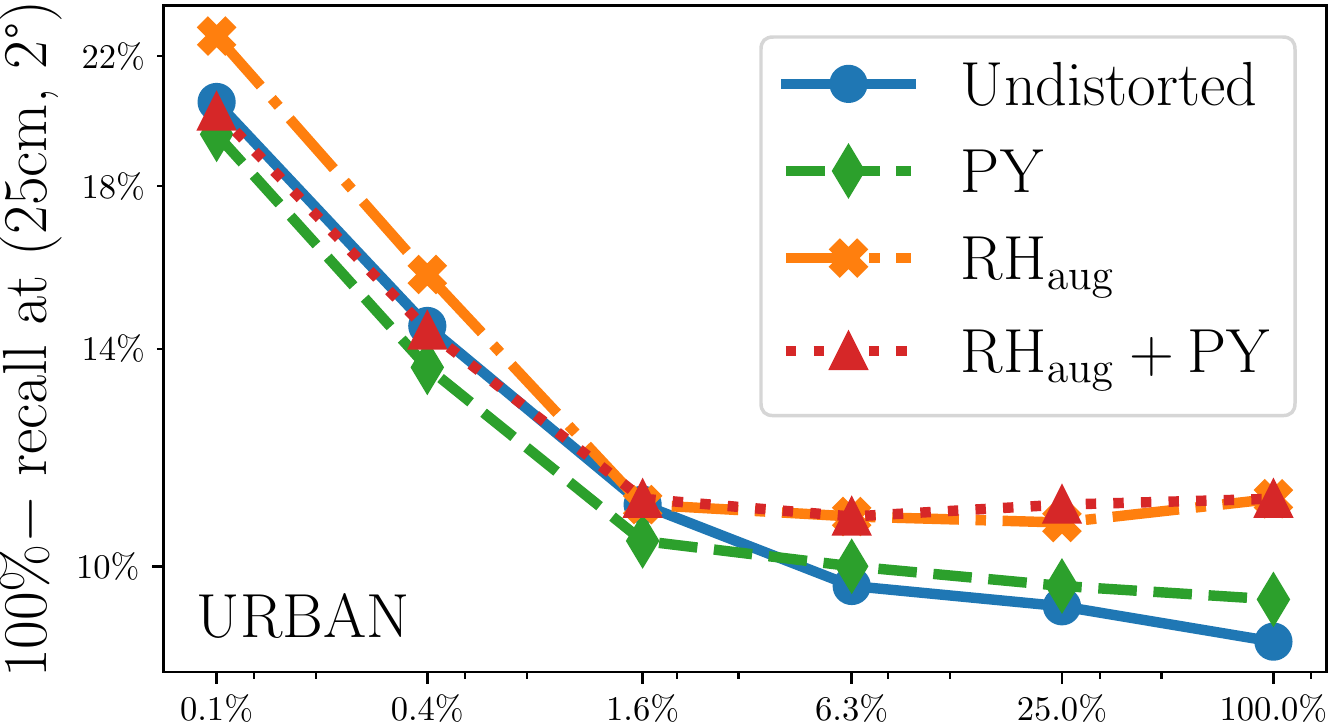}
    \includegraphics[width=\textwidth]{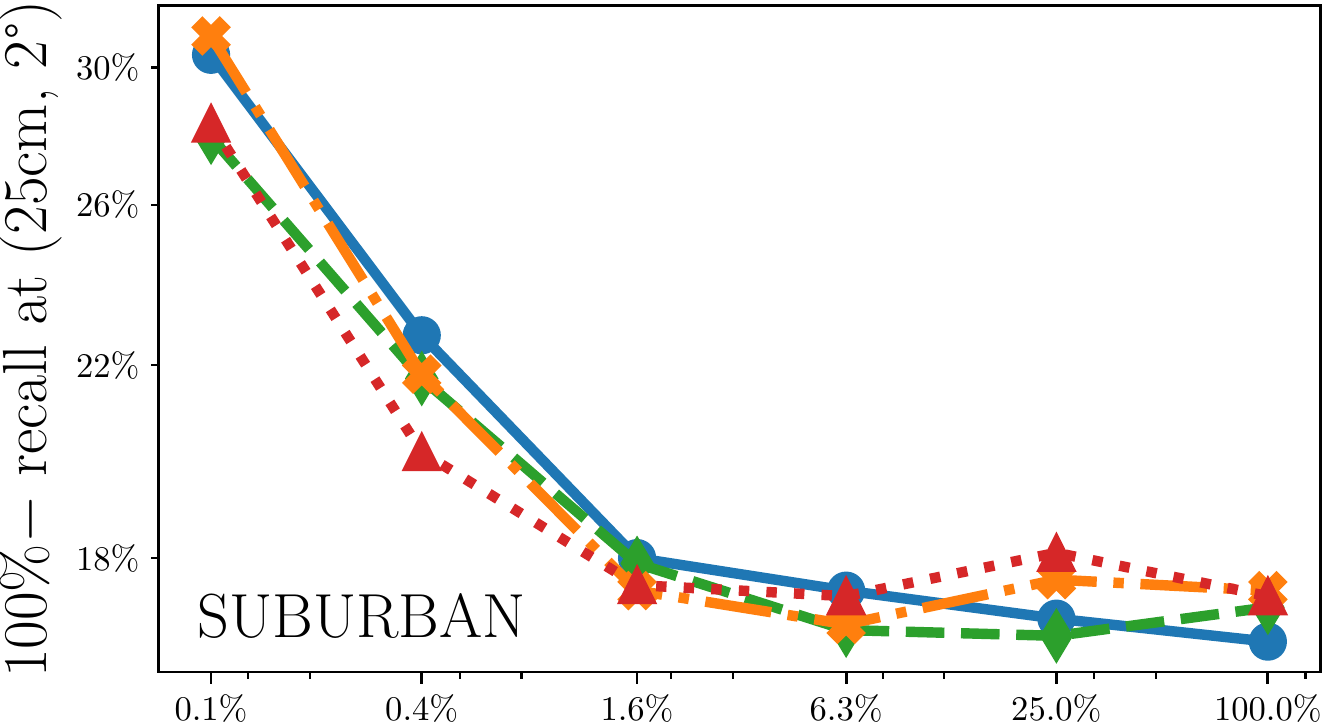}
    \includegraphics[width=\textwidth]{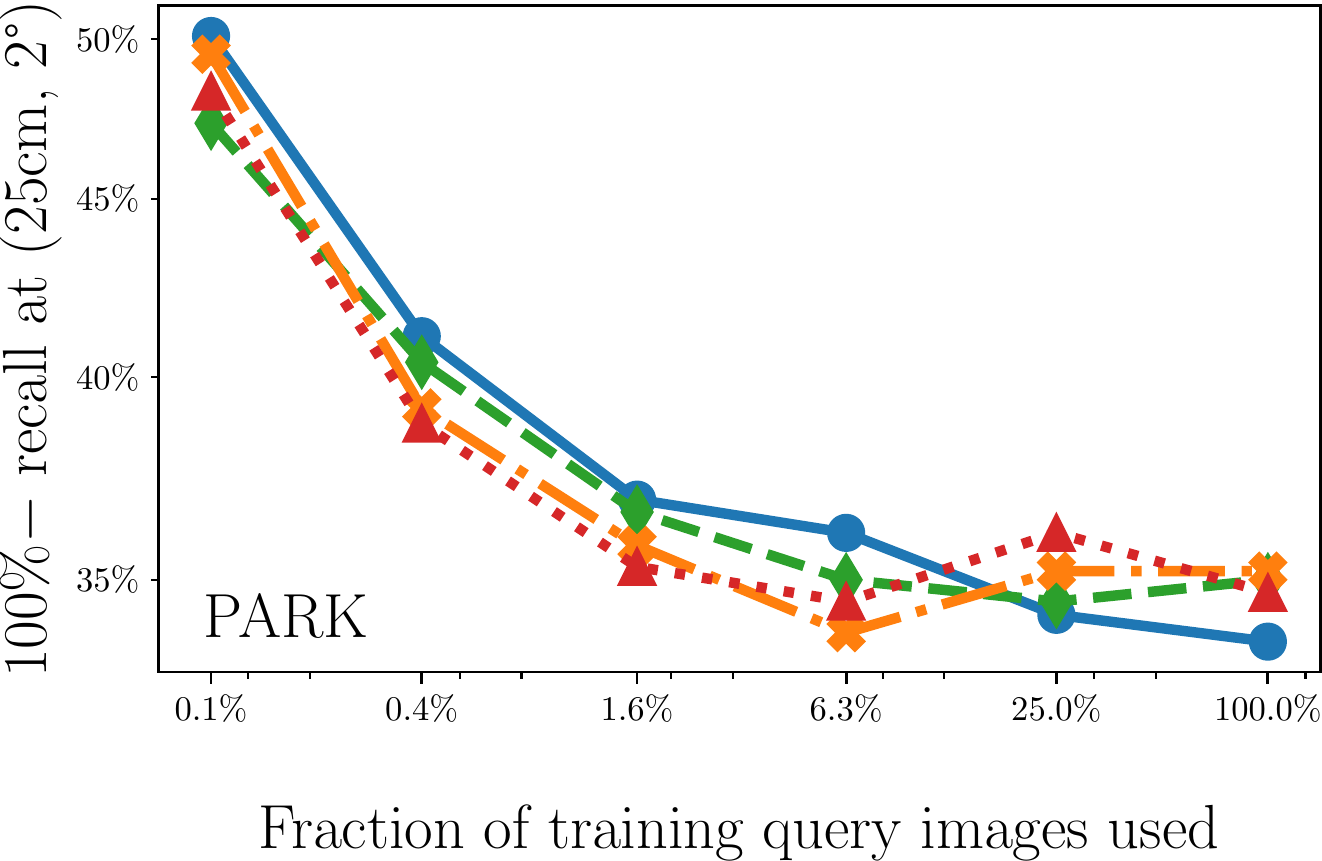}
   \end{minipage}
    \caption{Results for the Pixloc experiments on the urban, suburban and park subsets of the CMU Extended Seasons test data presented in two different ways. Left: detailed presentation of the test recall (higher is better). Right: log-log plot of the recall error (lower is better).}
    \label{fig:pixloc_cmu}
\end{figure}

We use four models in our experiments: A model trained on $\PY$-warped images, a model trained on undistorted images, a version with small $\Hm$-augmentations on undistorted images as well as a version with $\Hm$-augmentations on undistorted images that are afterwards warped to the $\PY$-domain. The last experiment is dubbed $\Hm_\mathrm{aug}+\PY$. We measure the performance by the recall of the model at an error threshold of $25\si{cm}$ for the translation and $2\si{\degree}$ for the rotation. 
The results are presented in Figure~\ref{fig:pixloc_cmu}. Similarly as in the \LM{} experiments, we plot the error rate for the different methods, i.e. $100\%-$ recall. The results do not give a conclusive answer to which setup performs the best across all data amounts and sequences. We can however see some trends.

\paragraph{Differences on data set parts.} There is a tendency for the two augmented methods to work better on the park dataset, and worse on the urban dataset, with the suburban dataset somewhere in the middle. That the success for data augmentation is dependent on the test sequence is plausible -- if the sequence is more diverse, more data should be needed, whence data augmentations gives a more significant boost. The fact that the park set is the most difficult of the three further supports this argument. On the CMU Extended Seasons dataset the initial pose in Pixloc obtained from image retrieval is commonly quite close in rotation to the correct pose, because the images are often taken along straight roads. This also partially explains why data augmentation with altered rotations is not very successful for the urban and suburban data subsets.

\paragraph{$\PY$ excels in low data regime.} More interestingly for us, we see that when we compare the $\PY$-warped to the undistorted method, the warped methods consistently perform better for smaller data amounts:  Out of the $9$ possible comparisons up to fraction $4^{-3}$ of the data, the warped version is better each time. Interestingly, the same is true for the augmented methods: here, the model trained on augmented undistorted data performs worse than the model trained on warped data in $7$ out of the $9$ experiments for data fraction less than or equal to $4^{-3}$. This to some extent supports the hypothesis that when data is scarce, the $\PY$-equivariance of the warped model helps the network generalize.

The saturation of the results around data fraction $4^{-2}$ can potentially be explained by the huge amount of data. At this fraction there are around 12k query-ref image pairs available to the model, and they are each seen on average only around ten times during the duration of the training procedure. 
We can speculate that in the lower data region before the saturation point, the errors again follow a power law distribution as discussed in Section~\ref{sec:exp} for the EfficientPose experiments, but more experiments would be needed to confirm this.

\begin{figure}
    \centering
    \includegraphics[height=12cm]{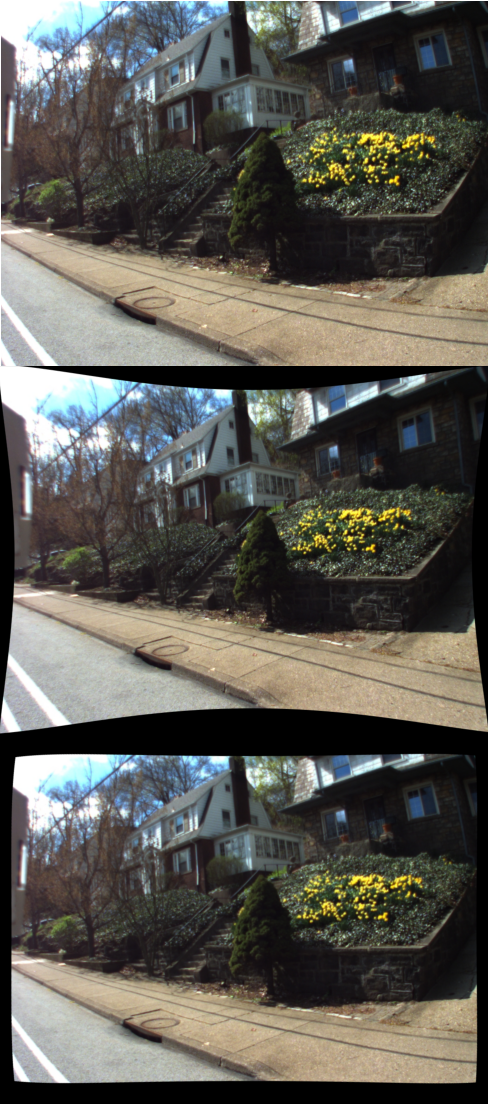}
    \caption{Example of the warping procedure on the CMU Extended Seasons data. Top: original image, middle: undistorted version, bottom: $\PY$-warped version. This dataset is a special case among our experiments, where the original images are already heavily radially distorted and hence similar to the $\PY$-warped versions.}
    \label{fig:warp_cmu}
\end{figure}

\paragraph{Disclaimer: Numerical issues.} As was previously stated, we are building a model using an updated version of the Pixloc code. Therefore, we reran the updated code without any of our modifications to use as a baseline. Unfortunately, all such runs except the one using the entire dataset and the one using a data fraction of $4^{-5}$ suffered from numerical problems and halted prematurely with NaN-values in the weights and losses, despite using the same hyperparameters as the other runs. We are unsure why this happened - we speculate that it might have to with differences in Pytorch or CUDA version from the authors and ``unlucky'' sampling of training batches, but ultimately, we do not have a satisfying explanation. Other users have reported similar issues to the github project.
For the runs that did not halt prematurely, we got the following results: When training on the entire dataset, the baseline obtained a recall of  $89.7\%$, $82.1\%$ and $64.7\%$ on the urban, suburban and park test sets, respectively, which is better than $88.3\%$,  $79.6 \%$  and $61.0\%$ for the old model (not the up-to-date code on github) trained on the entire dataset as reported in \cite{sarlin2021feature}.  The run only using a fraction of $4^{-5}$ of the data instead obtained a recall of $80.8\%$, $70.7\%$ and  $51.7\%$, respectively. The baseline results are quite similar to the results obtained by the warped model, achieving $90.5\%$, $82.9\%$, $65.0\%$ on the full data, and $80.5\%$, $72.1\%$, $52.7\%$ for the low data amount. This is expected since, as mentioned earlier, the radial distortion present in the original images is quite close to our $\PY$-model, which can be seen visually in Figure~\ref{fig:warp_cmu}. We therefore expect that the similarity between the $\PY$-runs and baseline would be high for all other data amounts as well had the baseline runs not experienced numerical issues and halted. 

\end{document}